\pdfoutput=1

\documentclass[11pt]{article}

\usepackage[final]{acl}

\usepackage{times}
\usepackage{latexsym}

\usepackage[T1]{fontenc}

\usepackage[utf8]{inputenc}

\usepackage{microtype}

\usepackage{inconsolata}

\usepackage{graphicx}

\usepackage{color, xcolor}

\usepackage{booktabs}
\usepackage{multicol}
\usepackage{multirow, makecell, caption}
\usepackage{colortbl}
\usepackage{tikz}
\usepackage{arydshln}
\usepackage{pgfplots}
\usepackage{amsmath}
\usepackage{amssymb}

\usepackage{tabularx}
\usepackage{enumerate}

\usepackage{CJKutf8}

\makeatletter
\def\adl@drawiv#1#2#3{%
        \hskip.5\tabcolsep
        \xleaders#3{#2.5\@tempdimb #1{1}#2.5\@tempdimb}%
                #2\z@ plus1fil minus1fil\relax
        \hskip.5\tabcolsep}
\newcommand{\cdashlinelr}[1]{%
  \noalign{\vskip\aboverulesep
           \global\let\@dashdrawstore\adl@draw
           \global\let\adl@draw\adl@drawiv}
  \cdashline{#1}
  \noalign{\global\let\adl@draw\@dashdrawstore
           \vskip\belowrulesep}}
\makeatother

\title{From Complex to Simple: Enhancing Multi-Constraint \\ 
Complex Instruction Following Ability of Large Language Models}

\author{
 \textbf{Qianyu He\textsuperscript{1}\thanks{\ Equal contribution.}},
 \textbf{Jie Zeng\textsuperscript{1}\footnotemark[1]},
 \textbf{Qianxi He\textsuperscript{1}},
 \textbf{Jiaqing Liang\textsuperscript{2}\thanks{\ Corresponding author.}},
 \textbf{Yanghua Xiao\textsuperscript{1,3}\footnotemark[2]}
\\
    \textsuperscript{\rm 1}Shanghai Key Laboratory of Data Science, School of Computer Science, Fudan University\\
    \textsuperscript{\rm 2}School of Data Science, Fudan University\\
     \textsuperscript{\rm 3}Fudan-Aishu Cognitive Intelligence Joint Research Center, Shanghai, China\\
     \{qyhe21, jzeng23, qxhe23\}@m.fudan.edu.cn, \{liangjiaqing, shawyh\}@fudan.edu.cn
}

\begin{document}
\thispagestyle{empty}
\maketitle

\begin{abstract}
It is imperative for Large language models (LLMs) to follow instructions with elaborate requirements (i.e. \textit{Complex Instructions Following}).
Yet, it remains under-explored how to enhance the ability of LLMs to follow complex instructions with multiple constraints.
To bridge the gap, we initially study \textit{what training data is effective} in enhancing complex constraints following abilities.
We found that training LLMs with instructions containing multiple constraints enhances their understanding of complex instructions, especially those with lower complexity levels.
Additionally, we further propose methods addressing how to \textit{obtain} and \textit{utilize} the effective training data.
Finally, we conduct extensive experiments to prove the effectiveness of our methods in terms of overall performance and training efficiency.
We also demonstrate that our methods improve models' ability to follow instructions generally and generalize effectively across out-of-domain, in-domain, and adversarial settings, while maintaining general capabilities.
The datasets and code are publicly available at \url{https://github.com/meowpass/FollowComplexInstruction}.



\end{abstract}

\section{Introduction}
\begin{figure}[t] 
    \centering
        \includegraphics[width=0.5\textwidth]{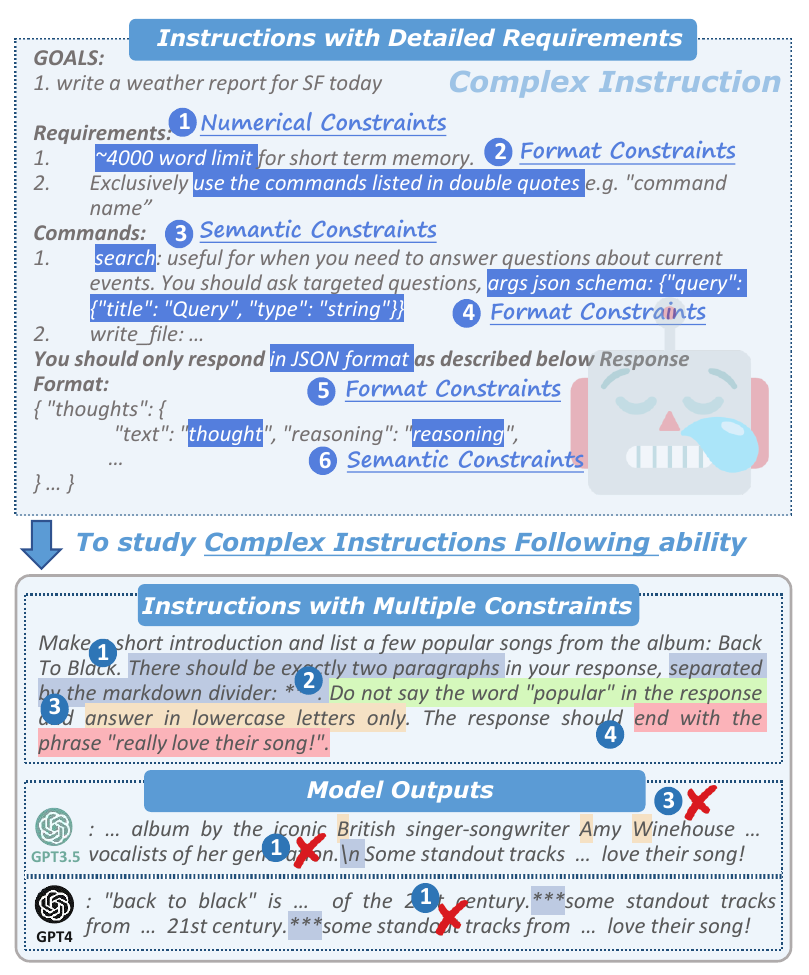}
    \caption{Real-world applications generally involve instructions with multiple constraints (i.e. \textit{Complex Instructions}), posing challenges for models. }
    \label{fig:010intro}
\end{figure}

Large language models (LLMs) have become the backbone for real-world applications~\cite{anil2023palm, touvron2023llama, achiam2023gpt}.
Given natural language instructions, LLMs can solve unseen tasks with few or no examples~\cite{brown2020language}.
The capability of LLMs to accurately understand instructions and convey the desired output, known as \textit{Instruction Following}~\cite{lou2024comprehensive}, is crucial for the safety~\cite{mu2023can} and reliability~\cite{zhou2023instruction} of LLMs.

\begin{figure*}[t] 
    \centering
            \includegraphics[width=1\textwidth]{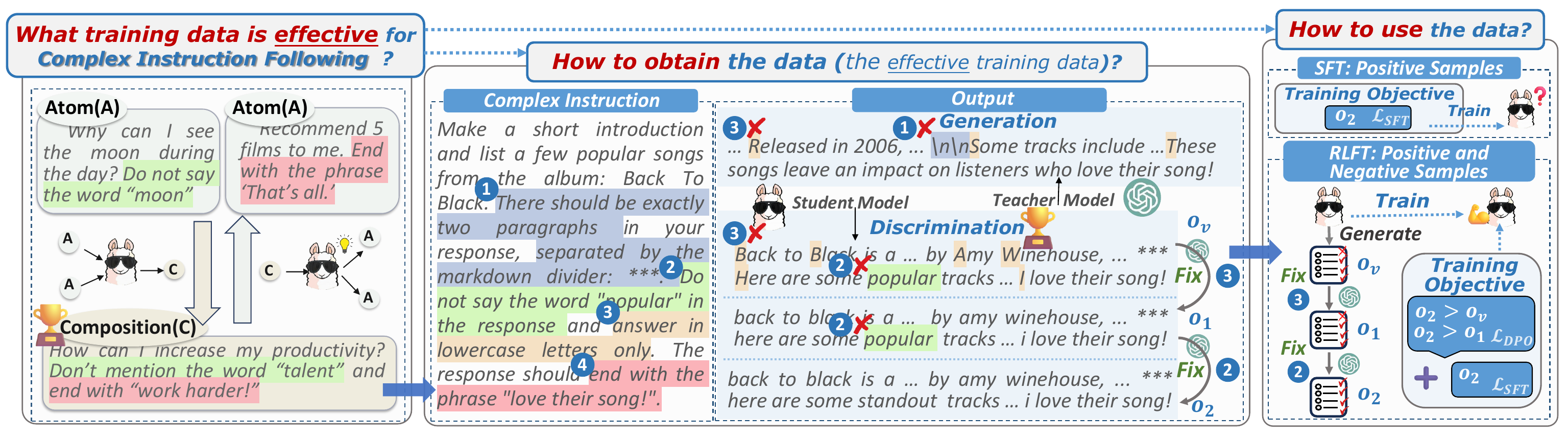}
    \caption{The framework of our study. We first study \textit{what training data is effective} in enhancing complex instruction following abilities via an empirical study.  
    Then, we design a discrimination-based method to address how to \textit{obtain} the data. 
    Finally, we propose a method for effectively \textit{utilizing} positive and negative samples obtained.}
    \label{fig:010method}
\end{figure*}

It is imperative for LLMs to follow instructions with elaborate requirements~\cite{yin2023llm, xu2023wizardlm} (i.e. \textit{Complex Instructions}), such as formatting specifications outlined in Fig. \ref{fig:010intro}.
On one hand, the ability to follow detailed instructions alleviates the need for annotating samples, which can be costly and challenging for intricate tasks~\cite{zeng2023agenttuning}.
On the other hand, complex instructions hardly appear in the training data~\cite{zhou2024lima}. 
Hence, the ability to follow complex instructions demonstrates models to have better generalization ability to unseen tasks~\cite{yin2023llm}.


Specifically, satisfying the multiple constraints in the instructions simultaneously (i.e. \textit{Constraints Following}) poses a significant challenge in complex instruction following~\cite{jiang2023followbench, he2024can}.
As shown in Fig.~\ref{fig:010intro}, whether models can satisfy the multiple constraints in the instructions determines their ability to follow complex instructions.
Hence, in our work, we explore complex instruction following by examining LLMs' ability to follow instructions with multiple constraints~\cite{yin2023llm, lou2024comprehensive}.
On one hand, human instructions are subjective and ambiguous, while constraints 
within these instructions facilitate the automatic evaluation of instruction following ability~\cite{zhou2023instruction, wang2024instructions}. 
On the other hand, the compositional nature of constraints enables the automatic creation of instructions with unseen compositions of constraints~\cite{zhou2023controlled, yao2023collie}. 
These instructions hardly appear in the training data, thus effectively assessing the model's ability to generalize to unseen tasks~\cite{aksu2023cesar}.


Complex constraints following is a challenging task for LLMs~\cite{jiang2023followbench, qin2024infobench}.
As shown in Fig.~\ref{fig:010intro}, even advanced LLMs struggle to meet the four specified constraints in complex instructions.
However, it remains under-explored \textit{how to enhance LLMs to follow multi-constraint complex instructions}.
First, the existing works on constraints following mainly focus on \textit{evaluation} without proposing methods for enhancement~\cite{chen2024benchmarking, xia2024fofo}.
Additionally, even when the improvement methods are proposed, they mainly consider instructions with \textit{few} constraints, thereby failing to showcase the complexity of human instructions in practical applications~\cite{chen2022controllable, zhang2023tell,  wang2024instructions}.
Moreover, although some studies construct complex instructions with multiple constraints and fine-tune LLMs on them~\cite{aksu2023cesar, sun2024conifer}, 
one key research question remains under-explored: \textbf{What \textit{training data} is effective in enhancing complex constraint-following abilities? }
This leads to two follow-up questions:
(1) \textbf{How to \textit{obtain} the effective training data?} and
(2) \textbf{How to \textit{utilize} the data effectively?}

In this work, we systematically study how to enhance the ability of LLMs to follow complex instructions, with the framework shown in Fig.~\ref{fig:010method}.
We initially explore the effective training data through an empirical study. 
We found that training LLMs on instructions containing multiple constraints (\textit{compositional data}) enhances their understanding of complex instructions more effectively than training on atomic constraints (\textit{atomic data}).
Moreover, the improvement in performance is related to the number of constraints, the model size.

To obtain high-quality compositional data, we generate initial output via a student model (vanilla model) and then correct it via a teacher model (advanced model), termed the Discrimination method.  
This approach yields higher-quality output than using the teacher model to generate directly.
To leverage the positive and negative samples collected during the Discrimination method, we introduce a contrastive method with reinforcement learning fine-tuning (RLFT)~\cite{rafailov2023direct}. 
Our method surpasses the SFT training paradigm on the instruction following benchmark~\cite{zhou2023instruction} with fewer training tokens. 
Overall, our methods enhance models' ability to follow instructions generally.
Our methods also generalize well across out-of-domain, in-domain, and adversarial settings while maintaining general capabilities.

Overall, our contributions are mainly three-fold:
(1) We systematically improve LLMs' instruction-following ability by exploring effective training data.
(2) We design a discrimination-based method to obtain effective training data. We also propose a method for utilizing positive and negative samples obtained through this approach.
(3) We conduct extensive experiments to prove the effectiveness and efficiency of our method. We also validate its generalization ability under five settings.

\section{Related Work}

\subsection{Instruction Following}


There are various ways to assess LLMs' ability to follow instructions. Some works study whether models understand the instructions by perturbing answer spaces~\cite{zeng2023evaluating, li2023instruction, wu2023reasoning}. 
other works incorporates verifiable constraints (such as lexical, numerical and format) within instructions~\cite{sun2023evaluating, jiang2023followbench, aksu2023cesar, zhou2023controlled, yao2023collie}. 
These constraints can be compositional, allowing one instruction to contain multiple constraints~\cite{aksu2023cesar, zhou2023controlled, yao2023collie}.
Such complex instructions pose greater challenges for LLMs to follow~\cite{he2024can, qin2024infobench}.
Our work falls into this latter category.
The existing works on constraints following either focus on evaluation~\cite{chen2024benchmarking, xia2024fofo} or consider instructions with few constraints~\cite{ zhang2023tell, chen2023comprehensive, wang2024instructions}.
Different from them, we systematically investigate how to enhance complex instructions with multiple constraints.

\subsection{Complex Instruction Tuning}
Complex Instructions can refer to instructions that involve more reasoning steps~\cite{mukherjee2023orca}, intricate input~\cite{zhou2024lima}, or multiple constraints~\cite{luo2023wizardmath}.
Many studies have demonstrated that fine-tuning with complex instructions can boost performance in tasks such as instruction following~\cite{xu2023wizardlm}, reasoning~\cite{mitra2023orca}, or code generation~\cite{luo2023wizardcoder}.
However, our work differs from these studies in two main aspects.
First, we focus on improving LLMs' ability to follow complex constraints, which is crucial for the practicality and safety of LLMs~\cite{zhou2023instruction, mu2023can}.
Furthermore, traditional supervised fine-tuning (SFT) uses only positive samples, whereas we use both positive and negative samples to enhance the complex instruction-following ability of LLMs effectively and efficiently.

\section{Empirical Studies}


\label{sec:emp}

A common approach to improve LLMs' ability to follow complex instructions is to construct corresponding instances and fine-tune the LLMs on them~\cite{aksu2023cesar, sun2024conifer}.
Yet, one key research question remains under-explored: \textbf{\textit{What training data is effective} in enhancing complex constraint-following abilities?}

\newcolumntype{b}{>{\columncolor{blue!10}}r}

\setlength\tabcolsep{6pt}
\begin{table*}[!htb]
  \centering
    \resizebox{0.8\textwidth}{!}{
\begin{tabular}{llcccccb}

\toprule

\textbf{Backbone}  &  \textbf{Methods}  & \textbf{Level 1} & \textbf{Level 2} & \textbf{Level 3} & \textbf{Level 4} & \textbf{Level 5} & \textbf{Avg.} \\

\midrule
\multirow{3}{*}{Vicuna-7B-V1.5\cite{zheng2024judging}}  & Backbone & 39.07 & \underline{44.71} & 37.28 & \textbf{30.93} & \textbf{19.06} & \underline{34.21} \\
                                 & Atom     & \underline{39.17} & 39.50 & \underline{42.07} & \underline{30.23} & \underline{16.97} & 33.59 \\
                                 & Comp     & \textbf{39.44} & \textbf{55.90} & \textbf{47.49} & 22.27 & 16.65 & \textbf{36.35} \\
\midrule

\multirow{3}{*}{LLaMA2-13B-Chat\cite{touvron2023llama}} & Backbone & 33.10 & \underline{41.71} & \underline{42.26} & \textbf{23.89} & \underline{22.07} & \underline{32.61} \\
                                 & Atom     & \textbf{38.99} & 39.78 & 36.61 & 20.74 & 14.83 & 30.19 \\
                                 & Comp     & \underline{37.02} & \textbf{44.66} & \textbf{42.55} & \underline{21.62} & \textbf{22.36} & \textbf{33.64} \\

\bottomrule
\end{tabular}
}
  \caption{
  The Instruction-level accuracy of models without further training (Backbone), training with atomic data (Atom), and compositional data (Comp) on FollowBench. Level $x$ indicates there are $x$ constraints in the instructions. Avg. indicates the average performance across 5 levels. The results are evaluated by GPT-4 using the FollowBench prompt template.
    The \textbf{bold} and \underline{underlined} represent the first and second rankings.
  }
  \label{tab:010followbench}
\end{table*}
\newcolumntype{d}{>{\columncolor{brown!10}}r}
\newcolumntype{q}{>{\columncolor{Green!10}}r}

\setlength\tabcolsep{4pt}
\begin{table*}[!htb]
  \centering
  \large
    \resizebox{1\textwidth}{!}{
\begin{tabular}{llcccccccccbd}

\toprule

\textbf{Backbone}  &  \textbf{Methods} & \textbf{ChangeCase} & \textbf{Combination} & \textbf{Content} & \textbf{Format} & \textbf{Keywords} & \textbf{Language} & \textbf{Length} & \textbf{Punctuation} & \textbf{Startend} & \textbf{I-level} & \textbf{C-level} \\

\midrule                    
\multirow{3}{*}{Vicuna-7B-V1.5}  & Backbone & 27.87 & 15.91 & \textbf{74.07} & 44.09 & \textbf{48.57} & \textbf{80.00} & 30.69 & 10.71 & \underline{40.00} & 26.89 & \underline{37.47} \\
                                 & Atom     & \underline{29.50} & \underline{31.82} & \underline{48.14} & \textbf{63.44} & \underline{36.19} & \underline{25.00} & \underline{31.68} & \textbf{16.07} & \underline{40.00} & \underline{27.17} & 37.29 \\
                                 & Comp     & \textbf{37.70} & \textbf{50.00} & 40.74 & \underline{55.91} & \underline{36.19} & \underline{25.00} & \textbf{32.67} & \underline{14.29} & \textbf{50.00} & \textbf{28.85} & \textbf{38.76} \\
\midrule 
\multirow{3}{*}{LLaMA2-13B-Chat} & Backbone & \textbf{42.62} & \textbf{11.36} & \textbf{81.48} & \textbf{55.91} & \textbf{45.71} & 15.00 & 32.67 & 00.00 & 25.00 & 25.77 & \underline{36.38} \\
                                 & Atom     & \textbf{42.62} & 00.00 & 37.04 & 54.84 & \underline{42.86} & \underline{35.00} & \underline{34.65} & \underline{12.50} & \underline{37.50} & \textbf{26.33} & 35.83 \\
                                 & Comp     & \underline{40.98} & \underline{02.27} & \underline{66.67} & \underline{54.84} & 38.10 & \textbf{50.00} & \textbf{36.63} & \textbf{16.07} & \textbf{40.00} & \underline{26.05} & \textbf{37.84} \\

\bottomrule
\end{tabular}
}
  \caption{ The performance of models without further training (Backbone), training with atomic data (Atom), and compositional data (Comp) on IFEval. The I-level and C-level denote the Instruction and Constraint-level accuracy.}
  \label{tab:010ifeval}
\end{table*}

Two types of training data can be utilized:
(1) Initially train models with atomic data, enabling them to handle compositional data automatically.
(2) Train models with compositional data, enabling them to understand instructions with atomic or varying compositions of constraints spontaneously.
Examples are shown in Fig.~\ref{fig:010method}.



To compare these training data types, we split instructions from existing benchmarks on instruction following~\cite{zhou2023instruction, jiang2023followbench} into training and test sets.
The training set includes atomic data (mostly with 1 constraint) and compositional data (mostly with over 3 constraints). Since original benchmarks lack corresponding outputs, we first generate them using GPT-3.5-turbo. To ensure quality, we further filter the training sets and retain only outputs that satisfy all instruction constraints by using GPT-3.5-turbo and predefined rules. The remaining data forms the test set$\footnote{Detailed data construction and statistics are provided in Appendix~\ref{sec:appendix_1}.}$. 

We compare three methods$\footnote{To prevent models from catastrophic forgetting~\cite{mccloskey1989catastrophic}, we mix training data with 10k ShareGPT data~\cite{chiang2023vicuna} for \textit{Atom} and \textit{Composition} checkpoint.
}$:
(1) \textit{Backbone}, the backbone model without further training.
(2) \textit{Atom} and (3) \textit{Composition}, continue training the backbone model with atomic data and compositional data respectively.
We leverage two backbone models~\cite{zheng2024judging, touvron2023llama} and adopt two accuracy metrics~\cite{zhou2023instruction, jiang2023followbench}:

\[
\text{acc}_{\text{ins}} = \frac{1}{m}\sum_{i=1}^{m}\prod_{j=1}^{n}c_i^j, \quad
\text{acc}_{\text{con}} = \frac{1}{mn}\sum_{i=1}^{m}\sum_{j=1}^{n}c_i^j,
\]
where $c_i^j$ equals 1 if the j-th constraint of the i-th instruction is satisfied, otherwise 0.
Overall, achieving Instruction-level accuracy ($\text{acc}_{\text{ins}}$) is more challenging than Constraint-level accuracy ($\text{acc}_{\text{con}}$).

The performance of the three methods on the test sets is shown in Tab.~\ref{tab:010followbench} and Tab.~\ref{tab:010ifeval}.
First, with regard to the overall performance, training with compositional data generally surpasses both the backbone model and atomic data training. 
This demonstrates that \textbf{training with \textit{compositional data} can generally enhance models' ability to follow complex instructions.}
Surprisingly, according to Tab.~\ref{tab:010followbench}, training with atomic data (mostly with 1 constraint) generally performs worse than the backbone model for instructions with more than 1 constraint.
Also, training with compositional data (usually 3 to 5 constraints) boosts performance on instructions with 1 to 3 constraints significantly but shows less enhancement or even a decline for those with 4 to 5 constraints.
This suggests that training with compositional data can better generalize to \textit{lower-level} complex instructions (instructions with fewer constraints).
Moreover, this effect is more pronounced in smaller LLMs (7B), likely due to their weaker generalization ability~\cite{fu2023specializing}.




\setlength\tabcolsep{1pt}
\begin{table*}[!htb]
  \centering
  \large
    \resizebox{0.95\textwidth}{!}{
\begin{tabular}{lcccccccccbd}

\toprule

\textbf{Methods} & \textbf{ChangeCase} & \textbf{Combination} & \textbf{Content} & \textbf{Format} & \textbf{Keywords} & \textbf{Language} & \textbf{Length} & \textbf{Punctuation} & \textbf{Startend} & \textbf{I-level} & \textbf{C-level} \\

\midrule                    
    
Vanilla            &21.19 & 08.89 & \textbf{77.26} & 56.67 & 61.60 & 10.60 & 30.85 & 00.26  & 16.84 & 06.40 & 41.33 \\
Generation &\underline{56.50} & \textbf{30.37} & 68.95 & \textbf{74.96} & \underline{72.29} & \underline{33.01} & \textbf{52.91} & \underline{36.76} & \underline{79.51} & \underline{21.53} & \underline{62.68} \\
Discrimination    &\textbf{66.56} & \underline{25.00} & \underline{68.11} & \underline{68.27} & \textbf{77.32} & \textbf{81.95} & \underline{52.27} & \textbf{70.90} & \textbf{85.60} & \textbf{35.04} & \textbf{68.30} \\

\bottomrule
\end{tabular}
}
  \caption{The output quality evaluated by IFEval across different methods.}
  \label{tab:010quality}
\end{table*}
We have found that training with compositional data can better enhance LLM's ability to follow complex instructions compared with atomic data.
A follow-up research question is \textbf{how to \textit{obtain} high-quality compositional data?}
Existing datasets either only provide compositional instructions without output~\cite{zhou2023instruction, jiang2023followbench} or directly generate responses using advanced LLMs and refine them manually~\cite{sun2024conifer}. 

We compare the outputs generated by three methods:
(1) \textit{Vanilla}: Output generated directly using backbone model.
(2) \textit{Generation}: Output generated directly using GPT-3.5-turbo.
(3) \textit{Discrimination}: First, we identify the constraints that Vanilla outputs failed to adhere to using test scripts~\cite{zhou2023instruction}.
Then, we rectify the Vanilla outputs constraints by constraints using GPT-3.5-turbo~(The framework is shown in Fig.~\ref{fig:010method} and please refer to \S\ref{sec:correction} for details).
With regard to the complex instructions, the instructions in IFEval~\cite{zhou2023instruction} originally have only 1 to 3 constraints, which are not complex enough. 
We construct 1467 complex instructions, each comprising 3 to 5 constraints that can be automatically verified (Please refer to \S\ref{sec:data} for details).
We leverage LLaMA2-13B-chat~\cite{touvron2023llama} as the backbone and assess output quality using the test script from ~\citet{zhou2023instruction}.

As shown in Tab.~\ref{tab:010quality}, using the generation method, outputs from advanced LLMs (Generation) are of higher quality than those from weaker LLMs (Vanilla).
However, \textbf{the outputs from weaker LLMs then refined by advanced LLMs (\textit{Discrimination}) are of better quality than the outputs generated by advanced LLMs directly (\textit{Generation})}.
This is because slight changes in the instruction (i.e. constraint) can cause substantial output differences, which the discrimination method captures better than the generation method.

\section{Method}


In \S\ref{sec:emp}, we propose a discrimination-based method to \textit{obtain} effective training data.
A subsequent question is \textbf{how to effectively \textit{utilize} the obtained data?}
To address this, we introduce a method that leverages both positive and negative samples to improve complex instruction following.
The framework is shown in Fig.~\ref{fig:010method}.


\subsection{Complex Instruction Synthesis} \label{sec:data}
According to \S\ref{sec:emp}, the effective training data is complex instructions with multiple constraints (compositional data).
To obtain compositional data, we first collect seed instructions from three widely used instruction-tuning datasets.
Then, we rewrite the instructions to incorporate multiple constraints.

To ensure the \textit{coverage} and \textit{diversity} of the seed instructions, we consider three sources:
(1) \textit{Open Assistant}~\cite{kopf2024openassistant}: human-written instructions when interacting with chatbots. We only consider rank 0 instructions (annotated by humans as the highest quality) and the first turn of the conversation~\cite{li2023self}.
(2) \textit{Self-Instruct}~\cite{wang2022self}: 175 manually crafted instructions covering various topics to aid instruction generation for new tasks.
(3) \textit{Super-Natural}~\cite{wang2022supernaturalinstructions}: NLP tasks formatted with human instructions. Tasks with simple outputs (e.g., classification, tagging) are excluded, leaving 318 tasks. One instruction is randomly selected from each remaining task.
From these sources, we gather a total of 1467 seed instructions.


Subsequently, we integrate constraints into these seed instructions. 
Initially, we randomly sample 3 to 5 constraints and use provided scripts to remove any conflicting constraints among those provided by~\citet{zhou2023instruction}.
Next, semantically equivalent but textually distinct instructions can substantially affect model outcomes~\cite{yan2024contrastive, chen2024benchmarking}.
Hence, we employ eight diverse expressions to describe each type of constraint.
Specifically, we manually select three common descriptions from the test set as seed descriptions, generate five similar descriptions using GPT-3.5-turbo, and refine them manually.
For each sampled constraint $c$, we randomly select one description $d_i$ from the description pool and append it to the instructions, formulated as:
\[
I_c = \text{LLM}(I_s \oplus d_i \oplus ... \oplus d_n),
\]
where $I_s$, $I_c$ and $d_i$ denote the seed instruction, its corresponding synthesized complex instruction, and appended constraint using a specific description.
The number of constraints $n$ ranges from 1 to 5, with a majority falling between 3 to 5$\footnote{The detailed statistics are shown in Tab.~\ref{tab:100quality_data}.}$.


\subsection{Teacher Correction} \label{sec:correction}
In \S\ref{sec:emp}, we propose a discrimination-based approach for obtaining the output, shown to be more effective than directly generating output with advanced LLMs. 
The details of this approach are as follows.

Initially, we utilize LLaMA2-13B-Chat~\cite{touvron2023llama} (student model) to generate results for our synthesized complex instructions.
Then, we utilize the test scripts from~\citet{zhou2023instruction} to identify the constraints the model failed to follow since the constraints are objective and automatically verifiable.
Finally, we adopt advanced LLMs (teacher model) GPT-3.5-turbo to correct the failed constraints one by one.

Specifically, each complex instruction $I_c$ contains multiple constraints. 
In \S\ref{sec:correction}, we utilize the test script to pinpoint the $f$ constraints $\mathcal{C} = \{c_1, c_2, ..., c_f\}$ that the student model's vanilla output $o_v$ fails to follow.
The teacher model sequentially corrects these failed constraints, yielding an output set $\mathcal{O} = \{o_v, o_1, o_2, ..., o_{f}\}$:
\[
o_1 = \text{LLM}(o_v, c_1),  \ldots , o_f = \text{LLM}(o_{f-1}, c_f),
\]
where GPT-3.5-turbo is employed as the teacher model with prompts sourced from Tab.~\ref{tab:100correct}.

\subsection{Contrastive Method} \label{sec:contrastive}
During \S\ref{sec:correction}, for each instruction $I_c$, we can gather positive sample set $\{o_f\}$ and negative samples set $\{o_v, o_1, ..., o_{f-1}\}$. Supervised fine-tuning (SFT) solely utilizes positive samples successfully meeting constraints specified in complex instructions~\cite{radford2019language, howard2018universal}.
However, negative samples from \S\ref{sec:correction}, failing to meet certain constraints, also offer valuable supervision signals.
Hence, we leverage the positive and negative samples through reinforcement learning fine-tuning~\cite{rafailov2023direct}.

Specifically, given the output set \( \mathcal{O} = \{o_v, o_1, o_2, ..., o_{f}\} \) for each complex instruction \( I_c \), we form a training dataset \( \mathcal{D} \) consisting of \( f \) contrastive triplets: \( \mathcal{D} = \{(I_c, o_v, o_f), (I_c, o_1, o_f), ..., (I_c, o_{f-1}, o_f) \} \). In each triplet, the final corrected output \( o_f \) (positive sample) is preferred over \( o_{i} \) (negative sample) as \( o_f \) satisfies more constraints specified in the complex instruction \( I_c \).
To model this preference information, we apply Direct Preference Optimization (DPO) \cite{rafailov2023direct}. The loss function involves a maximum likelihood objective for the language model parameters \( \pi_\theta \):


\begin{small}
\begin{gather*}
\mathcal{L}_{\text{DPO}}(\pi_\theta;\pi_\text{ref}) = - \mathbb{E}_{(I_c, o_{f}, o_{i})\sim \mathcal{D}}[\text{log}\sigma (\beta \text{log}\frac{\pi_\theta(o_{f}|I_c)}{\pi_\text{ref}(o_{f}|I_c)} \\
- \beta \text{log}\frac{\pi_\theta(o_{i}|I_c)}{\pi_\text{ref}(o_{i}|I_c)})],
\end{gather*}
\end{small}
where the model parameter \(\pi_\text{ref}\) starts as \(\pi_\theta\) and remains fixed during training. \(\beta\) is a hyperparameter, and \(\sigma\) denotes the sigmoid function. \(\mathcal{L}_{\text{DPO}}\) maximizes the log probability of the preferred output \(o_f\) relative to the dispreferred output \(o_i\).


However, solely relying on $\mathcal{L}_{\text{DPO}}$ may lead to low probabilities for both chosen and rejected outputs, yet with a significant disparity between them$\footnote{We provide some cases in Appx. ~\ref{sec:ablation_dpo}.}$.
Therefore, we integrate the SFT loss $\mathcal{L}_{\text{SFT}}$ to constrain $\pi_\theta$ from deviating from the preferred data distribution~\cite{xu2024contrastive, hejna2023contrastive}:
\[
\mathcal{L}_{\text{SFT}}(\pi_\theta) = -\mathbb{E}_{(I_c, o_{f}) \sim \mathcal{D}} [\log \pi_\theta(o_f|I_c)].
\]
Finally, our training procedure is to optimize $\mathcal{L}_{\text{DPO}}$ and $\mathcal{L}_{\text{SFT}}$ jointly:
\[
\mathcal{L}_{\text{Ours}} = \mathcal{L}_{\text{DPO}} + \mathcal{L}_{\text{SFT}}.
\]


\section{Experiments}

We conduct experiments to verify the effectiveness of our method, focusing on overall performance, training efficiency, and generalization ability.
\subsection{Experiment Setup}
\paragraph*{Models.} 

Our baselines include popular open and closed-source LLMs. 
We especially select models that excel in complex instruction following~\cite{xu2023wizardlm} and those that perform well on current instruction following benchmarks~\cite{wang2023openchat}.
Within our framework (\S\ref{sec:data}), we compare three methods:
(1) \textbf{$\text{Ours}_\text{Generation}$} directly generates output with GPT-3.5-turbo and trains the backbone model via supervised fine-tuning (SFT).
(2) \textbf{$\text{Ours}_\text{Discrimination}$} generates output via the backbone model then refines with GPT-3.5-turbo (\S\ref{sec:correction}), and trains the backbone model via SFT.
(3) \textbf{$\text{Ours}_\text{Contrastive}$} utilizes our advanced DPO for training (\S\ref{sec:contrastive}).
Across all methods, the instructions in the training data are identical, differing only in output and training paradigms$\footnote{Continuous training may lead to catastrophic forgetting \cite{mccloskey1989catastrophic}. Hence, we employ a replay strategy mixing training data with 10k ShareGPT data \cite{chiang2023vicuna} to maintain general abilities during training.}$.




\paragraph*{Evaluation.} 
We evaluate all models on IFEval~\cite{zhou2023instruction}, a widely-used instruction-following benchmark.
The test set consists of 541 samples, each containing 1 to 3 constraints.
The metrics are the same as \S\ref{sec:emp}.


\subsection{Results}

\paragraph*{Overall Performance.} 
\newcolumntype{b}{>{\columncolor{blue!10}}r}
\newcolumntype{d}{>{\columncolor{brown!10}}r}

\setlength\tabcolsep{1pt}
\begin{table*}[t]
  \centering
  \scriptsize
    \resizebox{1\textwidth}{!}{
\begin{tabular}{lccccccccccdb}

\toprule

\textbf{Models} & \textbf{BaseModel} & \textbf{ChangeCase} & \textbf{Combination} & \textbf{Content} & \textbf{Format} & \textbf{Keywords} & \textbf{Language} & \textbf{Length} & \textbf{Punctuation} & \textbf{Startend} & \textbf{I-level} & \textbf{C-level} \\

\midrule
PaLM2-S* \cite{anil2023palm}               &PaLM  & N/A & N/A & N/A & N/A & N/A & N/A & N/A & N/A & N/A & 43.07 & 55.76 \\
GPT3.5-turbo & GPT & 58.43 & 70.77 & 88.68 & 88.54 & 71.17 & 98.35 & 53.85 & 18.18 & 76.12 & 58.96 & 68.47 \\
GPT4 \cite{achiam2023gpt} & GPT & 75.28 & 70.77 & 96.23 & 94.27 & 84.05 & 96.77 & 73.43 & 66.67 & 95.52 & 76.16 & 82.97 \\
\cdashlinelr{1-13}
ChatGLM3-6B \cite{du2021glm}                    &ChatGLM & 14.61  & 16.92   & 67.92  & 42.68  & 50.92  & 51.61  & 34.97  & 28.79   & 49.25 & 27.36 & 39.33 \\
Qwen-14B-Chat \cite{bai2023qwen}           &Qwen  & 57.30 & 23.08 & 75.47 & 57.96 & 58.28 & 83.87 & 33.57 & 21.21 & 68.66 & 37.89 & 51.08 \\
LLaMA2-7B-Chat \cite{touvron2023llama}          & LLaMA2 & 35.96 & 06.15 & 79.25 & 57.96 & 53.37 & 19.35 & 36.36 & 07.58 & 41.79 & 28.84 & 41.61 \\
LLaMA2-13B-Chat \cite{touvron2023llama}         &LLaMA2  & 37.08 & 07.69  & 83.02 & 60.51 & 57.06 & 25.81 & 37.76 & 00.00  & 29.85 & 29.94 & 42.21 \\
LLaMA2-70B-Chat \cite{touvron2023llama}         &LLaMA2  & 42.70 & 24.62 & 79.25 & 63.69 & \underline{68.71} & 16.13 & 39.86 & 12.12 & 62.69 & 38.45 & 50.36 \\
Vicuna-13B-V1.5 \cite{zheng2024judging}        &LLaMA2  & 56.18 & \underline{32.31} & 75.47 & 62.42 & 57.06 & \underline{93.55} & 42.66 & 16.67 & 64.18 & 42.33 & 53.48 \\
WizardLM-13B-V1.2 \cite{xu2023wizardlm}       &LLaMA2  & 49.44 & 16.92 & 75.47 & 67.52 & 66.26 & 83.87 & 46.85 & 15.15 & 64.18 & 43.07 & 54.56 \\
OpenChat-13B-V3.2 \cite{wang2023openchat}      &LLaMA2  & 49.44 & 26.15 & \textbf{88.68} & 68.15 & 66.26 & 87.10 & \underline{47.55} & 19.70 & 71.64 & 46.03 & 57.43 \\
Mistral-7B-Instruct-v0.2 \cite{jiang2023mistral}    & Mistral & 61.80 & 21.54 & \textbf{88.68} & 75.16 & \textbf{76.07} & 58.06 & \underline{50.35} & 16.67 & 74.63 & 51.02 & 61.03 \\
\cdashlinelr{1-13}
$\text{Ours-LLaMA2-7B}_\text{Generation}$  & LLaMA2 & 41.57 & 15.38 & 71.70 & 70.70 & 53.37 & 58.06 & 27.97 & 9.09 & 56.72 & 34.01 & 46.16  \\
$\text{Ours-LLaMA2-7B}_\text{Discrimination}$  & LLaMA2 & 49.44 & 06.15  & 77.36 & 64.97 & 53.99 & 74.19 & 34.27 & 07.58  & 73.13 & 38.82 & 48.56 \\
$\text{Ours-LLaMA2-13B}_\text{Generation}$     &LLaMA2  & 64.04 & 20.00 & 66.04 & 70.06 & 53.99 & 35.48 & 44.06 & 21.21 & 74.63 & 41.22 & 52.88 \\
$\text{Ours-LLaMA2-7B}_\text{Contrastive}$    & LLaMA2 & \underline{76.40} & 13.85 & 75.47 & 70.06 & 50.92 & 67.74 & 37.76 & 24.24 & 82.09 & 42.88 & 54.68 \\
$\text{Ours-LLaMA2-13B}_\text{Discrimination}$ &LLaMA2  & 60.67 & 06.15  & 79.25 & 64.97 & 60.12 & \textbf{96.77} & 43.36 & \underline{51.52} & 79.10 & 46.21 & 57.43 \\
$\text{Ours-LLaMA2-13B}_\text{Contrastive}$    &LLaMA2  & 65.17 & 10.77 & \underline{84.91} & 66.88 & 60.74 & \underline{93.55} & 47.55 & 43.94 & 86.57 & 48.24 & 59.71 \\
\cdashlinelr{1-13}
$\text{Ours-Mistral-7B}_\text{Generation}$ & Mistral & 73.03 & \textbf{47.69} & 66.04 & 78.34 & 57.06 & 58.06 & \textbf{51.05} & 39.39 & \textbf{89.55} & 52.13 & 62.83 \\
$\text{Ours-Mistral-7B}_\text{Discrimination}$ & Mistral & \textbf{79.78} & 16.92 & 71.70 & \underline{80.89} & 61.35 & \underline{93.55} & 44.76 & \textbf{59.09} & 85.07 & \underline{53.79} & \underline{64.27} \\
$\text{Ours-Mistral-7B}_\text{Contrastive}$    & Mistral & 68.54 & 30.77 & \underline{84.91} & \textbf{85.35} & \underline{68.71} & 80.65 & 46.15 & 30.30 & \underline{88.06} & \textbf{53.97} & \textbf{64.99} \\

\bottomrule
\end{tabular}
}
  \caption{The overall performance on IFEval (each with 1 to 3 constraints). 
  The asterisk (*) indicates that the results are directly sourced from IFEval.
  N/A denotes that IFEval does not provide the results for specific constraints.
  }
  \label{tab:050main}
\end{table*}

As shown in Tab.~\ref{tab:050main}, for the same backbone, $\text{Ours}_\text{Discrimination}$ consistently outperforms $\text{Ours}_\text{Generation}$, and $\text{Ours}_\text{Contrastive}$ outperforms $\text{Ours}_\text{Discrimination}$, which proves the effectiveness of our methods.
Next, using the same backbone model (LLaMA2), $\text{Ours-LLaMA2-13B}_\text{Generation}$ performs worse than many open-source models, even when the constraints in the test set have been seen during training.
This highlights the importance in obtaining high-quality output for complex instructions.


\begin{figure}[t] 
    \centering
        \includegraphics[width=1\linewidth]{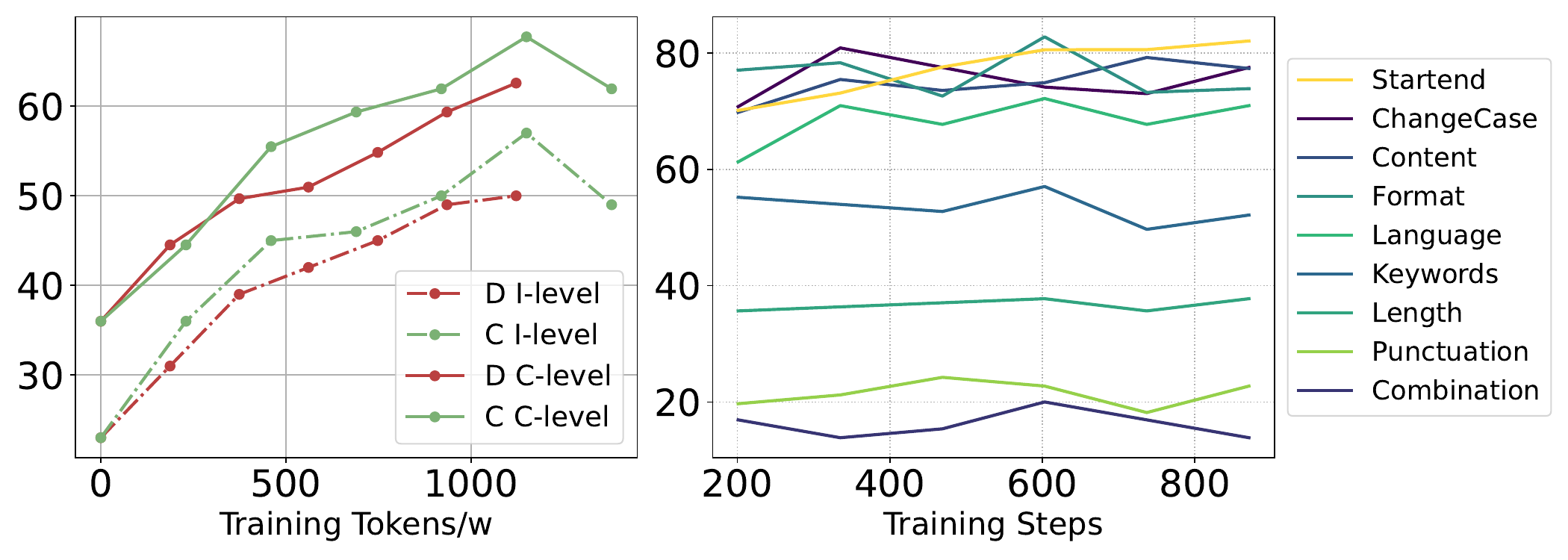}
    \caption{The performance of training efficiency (left) and each constraint type across different training tokens and training steps (right). D and C denote $\text{Ours-LLaMA2-13B}_\text{Discrimination}$ and $\text{Ours-LLaMA2-13B}_\text{Contrastive}$ respectively.
    }
    \label{fig:050_converge}
\end{figure}
\
\paragraph*{Training Efficiency.} 
To ensure fairness, we convert the checkpoints with the same number of training steps into the number of training tokens for the x-axis.
As shown in Fig.~\ref{fig:050_converge} (left), $\text{Ours-LLaMA2-13B}_\text{Contrastive}$ achieves better performance with the same training tokens and ultimately outperforms $\text{Ours-LLaMA2-13B}_\text{Discrimination}$.

\subsection{Analysis}
\newcolumntype{b}{>{\columncolor{blue!10}}r}
\newcolumntype{d}{>{\columncolor{brown!10}}r}

\setlength\tabcolsep{1pt}
\begin{table*}[t]
  \centering
  \scriptsize
    \resizebox{1\textwidth}{!}{
\begin{tabular}{lccccccccccdb}

\toprule

\textbf{Models} & \textbf{BaseModel} & \textbf{ChangeCase} & \textbf{Combination} & \textbf{Content} & \textbf{Format} & \textbf{Keywords} & \textbf{Language} & \textbf{Length} & \textbf{Punctuation} & \textbf{Startend} & \textbf{I-level} & \textbf{C-level} \\

\midrule
$\text{Ours-LLaMA2-13B}_\text{Discrimination-Random}$         & LLaMA2  & 58.43 & \underline{06.15} & 73.58 & 63.69 & 50.92 & \underline{87.10} & 33.57 & 10.61 & \underline{77.61} & 39.00 & 49.40 \\
$\text{Ours-LLaMA2-13B}_\text{Discrimination-Selected}$         & LLaMA2  & \textbf{71.91} & \textbf{15.38} & \underline{75.47} & \textbf{75.16} & \textbf{61.96} & 83.87 & \underline{37.06} & \underline{16.67} & 74.63 & \underline{44.55} & \underline{56.71} \\
$\text{Ours-LLaMA2-13B}_\text{Discrimination-All}$ &LLaMA2  & \underline{60.67} & \underline{06.15}  & \textbf{79.25} & \underline{64.97} & \underline{60.12} & \textbf{96.77} & \textbf{43.36} & \textbf{51.52} & \textbf{79.10} & \textbf{46.21} & \textbf{57.43} \\
\cdashlinelr{1-13}
$\text{Ours-LLaMA2-13B}_\text{Contrastive-Random}$     &LLaMA2  & 46.07 & \underline{10.77} & \underline{81.13} & \underline{68.79} & \textbf{64.42} & \underline{83.87} & \underline{39.86} & \textbf{50.00} & 79.10 & 44.55 & 56.71 \\
$\text{Ours-LLaMA2-13B}_\text{Contrastive-All}$    &LLaMA2  & \underline{65.17} & \underline{10.77} & \textbf{84.91} & 66.88 & \underline{60.74} & \textbf{93.55} & \textbf{47.55} & 43.94 & \textbf{86.57} & \underline{48.24} & \underline{59.71} \\
$\text{Ours-LLaMA2-13B}_\text{Contrastive-Selected}$         & LLaMA2  & \textbf{75.28} & \textbf{15.38} & 79.25 & \textbf{77.71} & 58.90  & 74.19 & 37.76 & \underline{45.45} & \underline{85.07} & \textbf{48.61} & \textbf{60.07} \\
\bottomrule
\end{tabular}
}
  \caption{The performance using the full set of noisy data samples (All), a subset of high-quality data samples (Selected), and a randomly sampled subset of the same size as the selected high-quality data subset (Random).}
  \label{tab:050quality}
\end{table*}
\subsubsection{Data Quality}\label{sec:dataquality}
Many studies have shown that data quality outweighs data quantity~\cite{zhou2024lima}.
As shown in Tab.~\ref{tab:010quality}, our training data contains noise.
Hence, we explore the performance using a subset of training data that has higher quality but lower quantity.
Specifically, we first filter the full set of noisy training data (containing 1467 samples) to retain only those samples whose output satisfies all the constraints in the instructions and finally obtain 515 high-quality data samples.
We also set a control group by randomly sampling a subset containing the same number of data samples as the selected high-quality data subset$\footnote{The statistics of the sampled data are shown in Tab.~\ref{tab:100quality_data}.}$.
We finally train the LLaMA2-13B-Chat with these sampled data. 
As shown in Tab.~\ref{tab:050quality}, training with full set of noisy data performs better than training with the subset of noisy data. 
Also, the selected high-quality data achieves comparable performance to training with the full set of noisy data.
This underscores the importance of selecting high-quality training data.

\subsubsection{Constraints Type}
We study the influence of constraint types on the effectiveness of the proposed methods.
First, to ensure the consistency of the model's output, we report the performance across different constraint types at the checkpoints taking different training steps.
As shown in Fig.~\ref{fig:050_converge} (right), certain constraints (e.g., Combination) are consistently more challenging to follow, while others (e.g., Startend) are easier. 
Hence, when synthesizing training data, certain challenging constraint types introduce more noise to our training data.
As shown in Tab.~\ref{tab:010quality}, the quality of training data differs across various constraints, leading to variations in the effectiveness of our methods across different types of constraints.

\subsection{Generalization Experiments} \label{050:generalization}
We investigate the generalizability of our framework from five perspectives.

\newcolumntype{b}{>{\columncolor{blue!10}}r}

\setlength\tabcolsep{7pt}
\begin{table}[t]
  \centering
    \resizebox{0.47\textwidth}{!}{
\begin{tabular}{lcccb}

\toprule

\textbf{Models}      & \textbf{\begin{tabular}[c]{@{}c@{}}LIMA\end{tabular} } & \textbf{\begin{tabular}[c]{@{}c@{}}Koala\end{tabular}} & \textbf{\begin{tabular}[c]{@{}c@{}}AlpacaEval\end{tabular}}& \textbf{Avg.} \\

\midrule                    

LLaMA2-13B-Chat      & 7.781                  & 7.619                        & 7.542                  & 7.647                  \\
$\text{Ours-LLaMA2-13B}_\text{Generation}$     &\underline{8.475}                  & 8.000                        & \underline{8.138}                  & 8.204                  \\
$\text{Ours-LLaMA2-13B}_\text{Contrastive}$   & 8.434                  & \underline{8.091}                        & 8.110                  & \underline{8.212}                  \\
$\text{Ours-LLaMA2-13B}_\text{Discrimination}$     & \textbf{8.552}                  & \textbf{8.097}                        & \textbf{8.204}                  & \textbf{8.284}                  \\

\bottomrule
\end{tabular}
}
  \caption{The performance of models on general instruction following datasets.
  }
  \label{tab:050generalIF}
\end{table}
\subsubsection{General Instruction Following Ability}
We evaluate models on three general instruction-following benchmarks, LIMA~\cite{zhou2024lima}, Koala~\cite{koala_blogpost_2023} and AlpacaEval~\cite{taori2023alpaca}$\footnote{The design of the scoring prompt is in Appx.~\ref{sec:appendix_general_instruction}}$.
They contain more general and diverse instructions than our training data.
As shown in Tab.~\ref{tab:050generalIF}, compared to the backbone model, training with complex instructions improve models' ability to follow instructions generally.

\subsubsection{Out-of-Domain Generalization} \label{sec:050ood}
\newcolumntype{b}{>{\columncolor{blue!10}}r}
\newcolumntype{d}{>{\columncolor{brown!10}}r}
\newcolumntype{q}{>{\columncolor{green!10}}r}

\setlength\tabcolsep{6pt}
\begin{table}[t]
  \centering
    \resizebox{0.43\textwidth}{!}{
\begin{tabular}{lddb}

\toprule

\multicolumn{1}{c}{\multirow{2}{*}{\textbf{Models}}} & \multicolumn{2}{c}{\textbf{Followbench}} & \multicolumn{1}{c}{\textbf{CELLO}} \\
\multicolumn{1}{c}{}                        & Mixed          & Total          & Total \\
\midrule
LLaMA2-13B-Chat                                           & 25.88          & 42.04          & 40.20 \\
$\text{Ours-LLaMA2-13B}_\text{Contrastive}$                                          & \underline{36.47}          & \textbf{42.77}          & \underline{44.20} \\
$\text{Ours-LLaMA2-13B}_\text{Discrimination}$                                           & \textbf{38.82}          & \underline{42.43}          & \textbf{56.10} \\

\bottomrule
\end{tabular}
}
  \caption{The performance of models on Followbench and CELLO. 
    }
  \label{tab:050ood}
\end{table}
We evaluate models on another two complex instruction-following benchmarks FollowBench~\cite{jiang2023followbench} and CELLO~\cite{he2024can}.
They have the following features to outline:
(1) They contain almost entirely \textit{different} constraints from IFEval$\footnote{The examples from these benchmarks are in Appx.~\ref{sec:appendix_out_domain}}$.
(2) To mirror real-world scenarios, FollowBench specifically introduces a \textit{Mixed} Category. Instructions within this category encompass multiple constraints of different types. 
(3) CELLO is a \textit{Chinese} complex instruction following benchmark mirroring real-world scenarios. The instructions are in a different language from our training data.
As shown in Tab.~\ref{tab:050ood}, our methods significantly enhance the ability of LLMs to follow different types of constraints, even when applied to different languages.
Interestingly, $\text{Ours}_\text{Contrastive}$ underperforms $\text{Ours}_\text{Discrimination}$ in some cases when applied to out-of-domain constraints, possibly due to DPO exhibiting lower generalization ability to out-of-preference data~\cite{li2023policy}. 


\newcolumntype{d}{>{\columncolor{brown!10}}r}
\newcolumntype{q}{>{\columncolor{Green!10}}r}

\setlength\tabcolsep{6pt}
\begin{table}[t]
  \centering
    \resizebox{0.48\textwidth}{!}{
\begin{tabular}{{lddbb}}

\toprule

\multicolumn{1}{c}{\multirow{2}{*}{\textbf{Models}}}                                & \multicolumn{2}{c}{\textbf{In-Domain}}& \multicolumn{2}{c}{\textbf{Adversarial}} \\
\multicolumn{1}{c}{}                                                 & I-level & C-level & I-level & C-level \\
\midrule                    
    
LLaMA2-13B-Chat                                 & 09.50     & 42.27     & 01.00       & 40.15 \\
WizardLM-13B-V1.2                               & 14.00     & 47.20     & 07.00       & 46.60\\
OpenChat-13B-V3.2                               & \underline{16.50}    & 49.07     & \underline{07.30}       & 47.64\\
\cdashlinelr{1-5}
$\text{Ours-LLaMA2-13B}_\text{Generation}$      & 14.00     & 52.27     & 05.00       & 49.36 \\
$\text{Ours-LLaMA2-13B}_\text{Discrimination}$  & 15.00     & \underline{53.33}     & 05.00       & \underline{49.53}\\
$\text{Ours-LLaMA2-13B}_\text{Contrastive}$     & \textbf{19.00}    & \textbf{55.73}     & \textbf{07.50}       & \textbf{53.05}\\

\bottomrule
\end{tabular}
}
  \caption{The performance of models on instructions with varying phrasing (In-Domain) and on more challenging complex instructions (Adversarial).
  }
  \label{tab:050all}
\end{table}
\subsubsection{In-Domain Generalization}~\label{sec:indomain}
We evaluate models on 200 new instructions, the constraints of which fall into the same categories as the training data but have different wording and specific requirements$\footnote{The construction process is detailed in the Appx.~\ref{sec:appendix_in_domain}}$.
As shown in Tab.~\ref{tab:050all}~(In-Domain), $\text{Ours}_\text{Contrastive}$ remains the top performer.
Also, the performance gap between $\text{Ours}_\text{Contrastive}$ and the best open-source model (OpenChat-13B-V3.2) has increased.

\subsubsection{Adversarial Setting}
We stress test the models on 200 more challenging complex instructions with increased constraints.
The instructions in the test set contain 6 to 7 constraints while our training data contains 3 to 5 constraints$\footnote{The construction process is detailed in the Appx.~\ref{sec:appendix_in_adver}}$.
As shown in Tab.~\ref{tab:050all}~(Adversarial), $\text{Ours}_\text{Contrastive}$ outperforms all other models and significantly performs better than $\text{Ours}_\text{Discrimination}$.

\newcolumntype{b}{>{\columncolor{blue!10}}r}

\setlength\tabcolsep{1pt}
\begin{table}[t]
  \centering
    \resizebox{0.48\textwidth}{!}{
\begin{tabular}{lccccb}

\toprule

\textbf{Models}      & \textbf{\begin{tabular}[c]{@{}c@{}}ARC\\ (25-shot)\end{tabular} } & \textbf{\begin{tabular}[c]{@{}c@{}}HellaSwag\\ (10-shot)\end{tabular}} & \textbf{\begin{tabular}[c]{@{}c@{}}MMLU\\ (5-shot)\end{tabular}}& \textbf{\begin{tabular}[c]{@{}c@{}}TruthfulQA\\ (0-shot)\end{tabular}}& \textbf{Avg.} \\

\midrule

LLaMA2-13B-Chat      & 59.04                  & 81.94                        & 54.64                  & 44.12                        & 59.94         \\
WizardLM-13B-V1.2    & 59.04                  & 82.21                        & 54.64                  & 47.27                        & 60.79         \\
OpenChat-13B-V3.2    & 59.64                  & 82.68                        & 56.68                  & 44.49                        & 60.87         \\
\cdashlinelr{1-6}
$\text{Ours-LLaMA2-13B}_\text{Discrimination}$  & 56.74                  & 78.39                        & 53.01                  & 48.17                        & 59.08         \\
$\text{Ours-LLaMA2-13B}_\text{Contrastive}$ & 57.76                  & 79.95                        & 53.79                  & 48.15                        & 59.91        \\

\bottomrule
\end{tabular}
}
  \caption{The performance of models on general tasks.
  }
  \label{tab:050general}
\end{table}
\subsubsection{General Ability}
We evaluate models on four widely-used benchmarks, reflecting knowledge capability (MMLU~\cite{hendrycks2020measuring}, TruthfulQA~\cite{lin2021truthfulqa}, ARC~\cite{clark2018think}), complex reasoning (HellaSwag~\cite{zellers2019hellaswag}).
As shown in Tab.~\ref{tab:050general}, our methods perform on par with open-source LLMs, validating that our methods maintain the models' general ability.


\section{Conclusion}
In this paper, we systematically study how to enhance the ability of LLMs to follow complex instructions. 
Initially, we study effective training data and methods for obtaining high-quality data through empirical studies.
Based on our findings, we introduce a method utilizing positive and negative samples to enhance LLMs' complex instruction-following ability. 
Our experiments show that our methods enhance models' ability to follow complex instructions more effectively and efficiently.
Finally, extensive experiments demonstrate the generalization abilities of our framework.

\section{Limitations}
We analyze the limitations of our work as follows.
First, we investigate complex instruction-following by testing LLMs' ability to adhere to instructions with multiple constraints. Even if the model meets all the constraints simultaneously, it may not fully follow complex instructions due to reasoning or knowledge limitations. However, we see complex constraint-following as a significant challenge worth studying.
In constructing the training data, we primarily use hard constraints from IFEval, although real-world scenarios often include soft constraints like semantic constraints. We focus on hard constraints because they can be objectively and automatically evaluated, and we believe that experiments based on them can yield valuable insights into complex instruction-following.

\bibliography{custom}

\begin{thebibliography}{58}
\providecommand{\natexlab}[1]{#1}

\bibitem[{Achiam et~al.(2023)Achiam, Adler, Agarwal, Ahmad, Akkaya, Aleman, Almeida, Altenschmidt, Altman, Anadkat et~al.}]{achiam2023gpt}
Josh Achiam, Steven Adler, Sandhini Agarwal, Lama Ahmad, Ilge Akkaya, Florencia~Leoni Aleman, Diogo Almeida, Janko Altenschmidt, Sam Altman, Shyamal Anadkat, et~al. 2023.
\newblock Gpt-4 technical report.
\newblock \emph{arXiv preprint arXiv:2303.08774}.

\bibitem[{Aksu et~al.(2023)Aksu, Hazarika, Mehri, Kim, Hakkani-Tur, Liu, and Namazifar}]{aksu2023cesar}
Taha Aksu, Devamanyu Hazarika, Shikib Mehri, Seokhwan Kim, Dilek Hakkani-Tur, Yang Liu, and Mahdi Namazifar. 2023.
\newblock Cesar: Automatic induction of compositional instructions for multi-turn dialogs.
\newblock In \emph{Proceedings of the 2023 Conference on Empirical Methods in Natural Language Processing}, pages 11709--11737.

\bibitem[{Anil et~al.(2023)Anil, Dai, Firat, Johnson, Lepikhin, Passos, Shakeri, Taropa, Bailey, Chen et~al.}]{anil2023palm}
Rohan Anil, Andrew~M Dai, Orhan Firat, Melvin Johnson, Dmitry Lepikhin, Alexandre Passos, Siamak Shakeri, Emanuel Taropa, Paige Bailey, Zhifeng Chen, et~al. 2023.
\newblock Palm 2 technical report.
\newblock \emph{arXiv preprint arXiv:2305.10403}.

\bibitem[{Bai et~al.(2023)Bai, Bai, Chu, Cui, Dang, Deng, Fan, Ge, Han, Huang et~al.}]{bai2023qwen}
Jinze Bai, Shuai Bai, Yunfei Chu, Zeyu Cui, Kai Dang, Xiaodong Deng, Yang Fan, Wenbin Ge, Yu~Han, Fei Huang, et~al. 2023.
\newblock Qwen technical report.
\newblock \emph{arXiv preprint arXiv:2309.16609}.

\bibitem[{Brown et~al.(2020)Brown, Mann, Ryder, Subbiah, Kaplan, Dhariwal, Neelakantan, Shyam, Sastry, Askell et~al.}]{brown2020language}
Tom Brown, Benjamin Mann, Nick Ryder, Melanie Subbiah, Jared~D Kaplan, Prafulla Dhariwal, Arvind Neelakantan, Pranav Shyam, Girish Sastry, Amanda Askell, et~al. 2020.
\newblock Language models are few-shot learners.
\newblock \emph{Advances in neural information processing systems}, 33:1877--1901.

\bibitem[{Chen et~al.(2022)Chen, Li, Chen, and Narasimhan}]{chen2022controllable}
Howard Chen, Huihan Li, Danqi Chen, and Karthik Narasimhan. 2022.
\newblock Controllable text generation with language constraints.
\newblock \emph{arXiv preprint arXiv:2212.10466}.

\bibitem[{Chen and Wan(2023)}]{chen2023comprehensive}
Xiang Chen and Xiaojun Wan. 2023.
\newblock A comprehensive evaluation of constrained text generation for large language models.
\newblock \emph{arXiv preprint arXiv:2310.16343}.

\bibitem[{Chen et~al.(2024)Chen, Xu, Wang, Liu, and Mao}]{chen2024benchmarking}
Yihan Chen, Benfeng Xu, Quan Wang, Yi~Liu, and Zhendong Mao. 2024.
\newblock Benchmarking large language models on controllable generation under diversified instructions.
\newblock \emph{arXiv preprint arXiv:2401.00690}.

\bibitem[{Chiang et~al.(2023)Chiang, Li, Lin, Sheng, Wu, Zhang, Zheng, Zhuang, Zhuang, Gonzalez et~al.}]{chiang2023vicuna}
Wei-Lin Chiang, Zhuohan Li, Zi~Lin, Ying Sheng, Zhanghao Wu, Hao Zhang, Lianmin Zheng, Siyuan Zhuang, Yonghao Zhuang, Joseph~E Gonzalez, et~al. 2023.
\newblock Vicuna: An open-source chatbot impressing gpt-4 with 90\%* chatgpt quality.
\newblock \emph{See https://vicuna. lmsys. org (accessed 14 April 2023)}, 2(3):6.

\bibitem[{Clark et~al.(2018)Clark, Cowhey, Etzioni, Khot, Sabharwal, Schoenick, and Tafjord}]{clark2018think}
Peter Clark, Isaac Cowhey, Oren Etzioni, Tushar Khot, Ashish Sabharwal, Carissa Schoenick, and Oyvind Tafjord. 2018.
\newblock Think you have solved question answering? try arc, the ai2 reasoning challenge.
\newblock \emph{arXiv preprint arXiv:1803.05457}.

\bibitem[{Du et~al.(2021)Du, Qian, Liu, Ding, Qiu, Yang, and Tang}]{du2021glm}
Zhengxiao Du, Yujie Qian, Xiao Liu, Ming Ding, Jiezhong Qiu, Zhilin Yang, and Jie Tang. 2021.
\newblock Glm: General language model pretraining with autoregressive blank infilling.
\newblock \emph{arXiv preprint arXiv:2103.10360}.

\bibitem[{Fu et~al.(2023)Fu, Peng, Ou, Sabharwal, and Khot}]{fu2023specializing}
Yao Fu, Hao Peng, Litu Ou, Ashish Sabharwal, and Tushar Khot. 2023.
\newblock Specializing smaller language models towards multi-step reasoning.
\newblock In \emph{International Conference on Machine Learning}, pages 10421--10430. PMLR.

\bibitem[{Geng et~al.(2023)Geng, Gudibande, Liu, Wallace, Abbeel, Levine, and Song}]{koala_blogpost_2023}
Xinyang Geng, Arnav Gudibande, Hao Liu, Eric Wallace, Pieter Abbeel, Sergey Levine, and Dawn Song. 2023.
\newblock \href {https://bair.berkeley.edu/blog/2023/04/03/koala/} {Koala: A dialogue model for academic research}.
\newblock Blog post.

\bibitem[{He et~al.(2024)He, Zeng, Huang, Chen, Xiao, He, Zhou, Liang, and Xiao}]{he2024can}
Qianyu He, Jie Zeng, Wenhao Huang, Lina Chen, Jin Xiao, Qianxi He, Xunzhe Zhou, Jiaqing Liang, and Yanghua Xiao. 2024.
\newblock Can large language models understand real-world complex instructions?
\newblock In \emph{Proceedings of the AAAI Conference on Artificial Intelligence}, volume~38, pages 18188--18196.

\bibitem[{Hejna et~al.(2023)Hejna, Rafailov, Sikchi, Finn, Niekum, Knox, and Sadigh}]{hejna2023contrastive}
Joey Hejna, Rafael Rafailov, Harshit Sikchi, Chelsea Finn, Scott Niekum, W~Bradley Knox, and Dorsa Sadigh. 2023.
\newblock Contrastive prefence learning: Learning from human feedback without rl.
\newblock \emph{arXiv preprint arXiv:2310.13639}.

\bibitem[{Hendrycks et~al.(2020)Hendrycks, Burns, Basart, Zou, Mazeika, Song, and Steinhardt}]{hendrycks2020measuring}
Dan Hendrycks, Collin Burns, Steven Basart, Andy Zou, Mantas Mazeika, Dawn Song, and Jacob Steinhardt. 2020.
\newblock Measuring massive multitask language understanding.
\newblock \emph{arXiv preprint arXiv:2009.03300}.

\bibitem[{Howard and Ruder(2018)}]{howard2018universal}
Jeremy Howard and Sebastian Ruder. 2018.
\newblock Universal language model fine-tuning for text classification.
\newblock \emph{arXiv preprint arXiv:1801.06146}.

\bibitem[{Hu et~al.(2021)Hu, Wallis, Allen-Zhu, Li, Wang, Wang, Chen et~al.}]{hu2021lora}
Edward~J Hu, Phillip Wallis, Zeyuan Allen-Zhu, Yuanzhi Li, Shean Wang, Lu~Wang, Weizhu Chen, et~al. 2021.
\newblock Lora: Low-rank adaptation of large language models.
\newblock In \emph{International Conference on Learning Representations}.

\bibitem[{Jiang et~al.(2023{\natexlab{a}})Jiang, Sablayrolles, Mensch, Bamford, Chaplot, Casas, Bressand, Lengyel, Lample, Saulnier et~al.}]{jiang2023mistral}
Albert~Q Jiang, Alexandre Sablayrolles, Arthur Mensch, Chris Bamford, Devendra~Singh Chaplot, Diego de~las Casas, Florian Bressand, Gianna Lengyel, Guillaume Lample, Lucile Saulnier, et~al. 2023{\natexlab{a}}.
\newblock Mistral 7b.
\newblock \emph{arXiv preprint arXiv:2310.06825}.

\bibitem[{Jiang et~al.(2023{\natexlab{b}})Jiang, Wang, Zeng, Zhong, Li, Mi, Shang, Jiang, Liu, and Wang}]{jiang2023followbench}
Yuxin Jiang, Yufei Wang, Xingshan Zeng, Wanjun Zhong, Liangyou Li, Fei Mi, Lifeng Shang, Xin Jiang, Qun Liu, and Wei Wang. 2023{\natexlab{b}}.
\newblock Followbench: A multi-level fine-grained constraints following benchmark for large language models.
\newblock \emph{arXiv preprint arXiv:2310.20410}.

\bibitem[{K{\"o}pf et~al.(2024)K{\"o}pf, Kilcher, von R{\"u}tte, Anagnostidis, Tam, Stevens, Barhoum, Nguyen, Stanley, Nagyfi et~al.}]{kopf2024openassistant}
Andreas K{\"o}pf, Yannic Kilcher, Dimitri von R{\"u}tte, Sotiris Anagnostidis, Zhi~Rui Tam, Keith Stevens, Abdullah Barhoum, Duc Nguyen, Oliver Stanley, Rich{\'a}rd Nagyfi, et~al. 2024.
\newblock Openassistant conversations-democratizing large language model alignment.
\newblock \emph{Advances in Neural Information Processing Systems}, 36.

\bibitem[{Li et~al.(2023{\natexlab{a}})Li, Yan, Wang, Tang, Ren, Srinivasan, and Jin}]{li2023instruction}
Shiyang Li, Jun Yan, Hai Wang, Zheng Tang, Xiang Ren, Vijay Srinivasan, and Hongxia Jin. 2023{\natexlab{a}}.
\newblock Instruction-following evaluation through verbalizer manipulation.
\newblock \emph{arXiv preprint arXiv:2307.10558}.

\bibitem[{Li et~al.(2023{\natexlab{b}})Li, Yu, Zhou, Schick, Zettlemoyer, Levy, Weston, and Lewis}]{li2023self}
Xian Li, Ping Yu, Chunting Zhou, Timo Schick, Luke Zettlemoyer, Omer Levy, Jason Weston, and Mike Lewis. 2023{\natexlab{b}}.
\newblock Self-alignment with instruction backtranslation.
\newblock \emph{arXiv preprint arXiv:2308.06259}.

\bibitem[{Li et~al.(2023{\natexlab{c}})Li, Xu, and Yu}]{li2023policy}
Ziniu Li, Tian Xu, and Yang Yu. 2023{\natexlab{c}}.
\newblock Policy optimization in rlhf: The impact of out-of-preference data.
\newblock \emph{arXiv preprint arXiv:2312.10584}.

\bibitem[{Lin et~al.(2021)Lin, Hilton, and Evans}]{lin2021truthfulqa}
Stephanie Lin, Jacob Hilton, and Owain Evans. 2021.
\newblock Truthfulqa: Measuring how models mimic human falsehoods.
\newblock \emph{arXiv preprint arXiv:2109.07958}.

\bibitem[{Lou et~al.(2024)Lou, Zhang, and Yin}]{lou2024comprehensive}
Renze Lou, Kai Zhang, and Wenpeng Yin. 2024.
\newblock \href {https://arxiv.org/abs/2303.10475} {A comprehensive survey on instruction following}.
\newblock \emph{Preprint}, arXiv:2303.10475.

\bibitem[{Luo et~al.(2023{\natexlab{a}})Luo, Sun, Xu, Zhao, Lou, Tao, Geng, Lin, Chen, and Zhang}]{luo2023wizardmath}
Haipeng Luo, Qingfeng Sun, Can Xu, Pu~Zhao, Jianguang Lou, Chongyang Tao, Xiubo Geng, Qingwei Lin, Shifeng Chen, and Dongmei Zhang. 2023{\natexlab{a}}.
\newblock Wizardmath: Empowering mathematical reasoning for large language models via reinforced evol-instruct.
\newblock \emph{arXiv preprint arXiv:2308.09583}.

\bibitem[{Luo et~al.(2023{\natexlab{b}})Luo, Xu, Zhao, Sun, Geng, Hu, Tao, Ma, Lin, and Jiang}]{luo2023wizardcoder}
Ziyang Luo, Can Xu, Pu~Zhao, Qingfeng Sun, Xiubo Geng, Wenxiang Hu, Chongyang Tao, Jing Ma, Qingwei Lin, and Daxin Jiang. 2023{\natexlab{b}}.
\newblock \href {https://arxiv.org/abs/2306.08568} {Wizardcoder: Empowering code large language models with evol-instruct}.
\newblock \emph{Preprint}, arXiv:2306.08568.

\bibitem[{McCloskey and Cohen(1989)}]{mccloskey1989catastrophic}
Michael McCloskey and Neal~J Cohen. 1989.
\newblock Catastrophic interference in connectionist networks: The sequential learning problem.
\newblock In \emph{Psychology of learning and motivation}, volume~24, pages 109--165. Elsevier.

\bibitem[{Mitra et~al.(2023)Mitra, Del~Corro, Mahajan, Codas, Simoes, Agarwal, Chen, Razdaibiedina, Jones, Aggarwal et~al.}]{mitra2023orca}
Arindam Mitra, Luciano Del~Corro, Shweti Mahajan, Andres Codas, Clarisse Simoes, Sahaj Agarwal, Xuxi Chen, Anastasia Razdaibiedina, Erik Jones, Kriti Aggarwal, et~al. 2023.
\newblock Orca 2: Teaching small language models how to reason.
\newblock \emph{arXiv preprint arXiv:2311.11045}.

\bibitem[{Mu et~al.(2023)Mu, Chen, Wang, Chen, Karamardian, Aljeraisy, Hendrycks, and Wagner}]{mu2023can}
Norman Mu, Sarah Chen, Zifan Wang, Sizhe Chen, David Karamardian, Lulwa Aljeraisy, Dan Hendrycks, and David Wagner. 2023.
\newblock Can llms follow simple rules?
\newblock \emph{arXiv preprint arXiv:2311.04235}.

\bibitem[{Mukherjee et~al.(2023)Mukherjee, Mitra, Jawahar, Agarwal, Palangi, and Awadallah}]{mukherjee2023orca}
Subhabrata Mukherjee, Arindam Mitra, Ganesh Jawahar, Sahaj Agarwal, Hamid Palangi, and Ahmed Awadallah. 2023.
\newblock Orca: Progressive learning from complex explanation traces of gpt-4.
\newblock \emph{arXiv preprint arXiv:2306.02707}.

\bibitem[{Qin et~al.(2024)Qin, Song, Hu, Yao, Cho, Wang, Wu, Liu, Liu, and Yu}]{qin2024infobench}
Yiwei Qin, Kaiqiang Song, Yebowen Hu, Wenlin Yao, Sangwoo Cho, Xiaoyang Wang, Xuansheng Wu, Fei Liu, Pengfei Liu, and Dong Yu. 2024.
\newblock Infobench: Evaluating instruction following ability in large language models.
\newblock \emph{arXiv preprint arXiv:2401.03601}.

\bibitem[{Radford et~al.(2019)Radford, Wu, Child, Luan, Amodei, Sutskever et~al.}]{radford2019language}
Alec Radford, Jeffrey Wu, Rewon Child, David Luan, Dario Amodei, Ilya Sutskever, et~al. 2019.
\newblock Language models are unsupervised multitask learners.
\newblock \emph{OpenAI blog}, 1(8):9.

\bibitem[{Rafailov et~al.(2023)Rafailov, Sharma, Mitchell, Ermon, Manning, and Finn}]{rafailov2023direct}
Rafael Rafailov, Archit Sharma, Eric Mitchell, Stefano Ermon, Christopher~D. Manning, and Chelsea Finn. 2023.
\newblock \href {https://arxiv.org/abs/2305.18290} {Direct preference optimization: Your language model is secretly a reward model}.
\newblock \emph{Preprint}, arXiv:2305.18290.

\bibitem[{Sun et~al.(2024)Sun, Liu, Li, Wang, Dong, Lin, and Huang}]{sun2024conifer}
Haoran Sun, Lixin Liu, Junjie Li, Fengyu Wang, Baohua Dong, Ran Lin, and Ruohui Huang. 2024.
\newblock Conifer: Improving complex constrained instruction-following ability of large language models.
\newblock \emph{arXiv preprint arXiv:2404.02823}.

\bibitem[{Sun et~al.(2023)Sun, Tian, Zhou, Xu, Hu, Gupta, Wieting, Peng, and Ma}]{sun2023evaluating}
Jiao Sun, Yufei Tian, Wangchunshu Zhou, Nan Xu, Qian Hu, Rahul Gupta, John Wieting, Nanyun Peng, and Xuezhe Ma. 2023.
\newblock Evaluating large language models on controlled generation tasks.
\newblock In \emph{Proceedings of the 2023 Conference on Empirical Methods in Natural Language Processing}, pages 3155--3168.

\bibitem[{Taori et~al.(2023)Taori, Gulrajani, Zhang, Dubois, Li, Guestrin, Liang, and Hashimoto}]{taori2023alpaca}
Rohan Taori, Ishaan Gulrajani, Tianyi Zhang, Yann Dubois, Xuechen Li, Carlos Guestrin, Percy Liang, and Tatsunori~B Hashimoto. 2023.
\newblock Alpaca: A strong, replicable instruction-following model.
\newblock \emph{Stanford Center for Research on Foundation Models. https://crfm. stanford. edu/2023/03/13/alpaca. html}, 3(6):7.

\bibitem[{Touvron et~al.(2023)Touvron, Martin, Stone, Albert, Almahairi, Babaei, Bashlykov, Batra, Bhargava, Bhosale et~al.}]{touvron2023llama}
Hugo Touvron, Louis Martin, Kevin Stone, Peter Albert, Amjad Almahairi, Yasmine Babaei, Nikolay Bashlykov, Soumya Batra, Prajjwal Bhargava, Shruti Bhosale, et~al. 2023.
\newblock Llama 2: Open foundation and fine-tuned chat models.
\newblock \emph{arXiv preprint arXiv:2307.09288}.

\bibitem[{Wang et~al.(2024)Wang, Shang, Jain, Wang, Ning, Min, Castelli, Benajiba, and Roth}]{wang2024instructions}
Fei Wang, Chao Shang, Sarthak Jain, Shuai Wang, Qiang Ning, Bonan Min, Vittorio Castelli, Yassine Benajiba, and Dan Roth. 2024.
\newblock From instructions to constraints: Language model alignment with automatic constraint verification.
\newblock \emph{arXiv preprint arXiv:2403.06326}.

\bibitem[{Wang et~al.(2023)Wang, Cheng, Zhan, Li, Song, and Liu}]{wang2023openchat}
Guan Wang, Sijie Cheng, Xianyuan Zhan, Xiangang Li, Sen Song, and Yang Liu. 2023.
\newblock Openchat: Advancing open-source language models with mixed-quality data.
\newblock \emph{arXiv preprint arXiv:2309.11235}.

\bibitem[{Wang et~al.(2022{\natexlab{a}})Wang, Kordi, Mishra, Liu, Smith, Khashabi, and Hajishirzi}]{wang2022self}
Yizhong Wang, Yeganeh Kordi, Swaroop Mishra, Alisa Liu, Noah~A Smith, Daniel Khashabi, and Hannaneh Hajishirzi. 2022{\natexlab{a}}.
\newblock Self-instruct: Aligning language models with self-generated instructions.
\newblock \emph{arXiv preprint arXiv:2212.10560}.

\bibitem[{Wang et~al.(2022{\natexlab{b}})Wang, Mishra, Alipoormolabashi, Kordi, Mirzaei, Arunkumar, Ashok, Dhanasekaran, Naik, Stap, Pathak, Karamanolakis, Lai, Purohit, Mondal, Anderson, Kuznia, Doshi, Patel, Pal, Moradshahi, Parmar, Purohit, Varshney, Kaza, Verma, Puri, Karia, Sampat, Doshi, Mishra, Reddy, Patro, Dixit, Shen, Baral, Choi, Smith, Hajishirzi, and Khashabi}]{wang2022supernaturalinstructions}
Yizhong Wang, Swaroop Mishra, Pegah Alipoormolabashi, Yeganeh Kordi, Amirreza Mirzaei, Anjana Arunkumar, Arjun Ashok, Arut~Selvan Dhanasekaran, Atharva Naik, David Stap, Eshaan Pathak, Giannis Karamanolakis, Haizhi~Gary Lai, Ishan Purohit, Ishani Mondal, Jacob Anderson, Kirby Kuznia, Krima Doshi, Maitreya Patel, Kuntal~Kumar Pal, Mehrad Moradshahi, Mihir Parmar, Mirali Purohit, Neeraj Varshney, Phani~Rohitha Kaza, Pulkit Verma, Ravsehaj~Singh Puri, Rushang Karia, Shailaja~Keyur Sampat, Savan Doshi, Siddhartha Mishra, Sujan Reddy, Sumanta Patro, Tanay Dixit, Xudong Shen, Chitta Baral, Yejin Choi, Noah~A. Smith, Hannaneh Hajishirzi, and Daniel Khashabi. 2022{\natexlab{b}}.
\newblock \href {https://arxiv.org/abs/2204.07705} {Super-naturalinstructions: Generalization via declarative instructions on 1600+ nlp tasks}.
\newblock \emph{Preprint}, arXiv:2204.07705.

\bibitem[{Wu et~al.(2023)Wu, Qiu, Ross, Aky{\"u}rek, Chen, Wang, Kim, Andreas, and Kim}]{wu2023reasoning}
Zhaofeng Wu, Linlu Qiu, Alexis Ross, Ekin Aky{\"u}rek, Boyuan Chen, Bailin Wang, Najoung Kim, Jacob Andreas, and Yoon Kim. 2023.
\newblock Reasoning or reciting? exploring the capabilities and limitations of language models through counterfactual tasks.
\newblock \emph{arXiv preprint arXiv:2307.02477}.

\bibitem[{Xia et~al.(2024)Xia, Xing, Du, Yang, Feng, Xu, Yin, and Xiong}]{xia2024fofo}
Congying Xia, Chen Xing, Jiangshu Du, Xinyi Yang, Yihao Feng, Ran Xu, Wenpeng Yin, and Caiming Xiong. 2024.
\newblock Fofo: A benchmark to evaluate llms' format-following capability.
\newblock \emph{arXiv preprint arXiv:2402.18667}.

\bibitem[{Xu et~al.(2023)Xu, Sun, Zheng, Geng, Zhao, Feng, Tao, and Jiang}]{xu2023wizardlm}
Can Xu, Qingfeng Sun, Kai Zheng, Xiubo Geng, Pu~Zhao, Jiazhan Feng, Chongyang Tao, and Daxin Jiang. 2023.
\newblock Wizardlm: Empowering large language models to follow complex instructions.
\newblock \emph{arXiv preprint arXiv:2304.12244}.

\bibitem[{Xu et~al.(2024)Xu, Sharaf, Chen, Tan, Shen, Van~Durme, Murray, and Kim}]{xu2024contrastive}
Haoran Xu, Amr Sharaf, Yunmo Chen, Weiting Tan, Lingfeng Shen, Benjamin Van~Durme, Kenton Murray, and Young~Jin Kim. 2024.
\newblock Contrastive preference optimization: Pushing the boundaries of llm performance in machine translation.
\newblock \emph{arXiv preprint arXiv:2401.08417}.

\bibitem[{Yan et~al.(2024)Yan, Wang, Huang, Zhou, Yin, Galstyan, Yin, and Chen}]{yan2024contrastive}
Tianyi Yan, Fei Wang, James~Y Huang, Wenxuan Zhou, Fan Yin, Aram Galstyan, Wenpeng Yin, and Muhao Chen. 2024.
\newblock Contrastive instruction tuning.
\newblock \emph{arXiv preprint arXiv:2402.11138}.

\bibitem[{Yao et~al.(2023)Yao, Chen, Hanjie, Yang, and Narasimhan}]{yao2023collie}
Shunyu Yao, Howard Chen, Austin~W Hanjie, Runzhe Yang, and Karthik Narasimhan. 2023.
\newblock Collie: Systematic construction of constrained text generation tasks.
\newblock \emph{arXiv preprint arXiv:2307.08689}.

\bibitem[{Yin et~al.(2023)Yin, Ye, Liu, Ren, and Sch{\"u}tze}]{yin2023llm}
Wenpeng Yin, Qinyuan Ye, Pengfei Liu, Xiang Ren, and Hinrich Sch{\"u}tze. 2023.
\newblock Llm-driven instruction following: Progresses and concerns.
\newblock In \emph{Proceedings of the 2023 Conference on Empirical Methods in Natural Language Processing: Tutorial Abstracts}, pages 19--25.

\bibitem[{Zellers et~al.(2019)Zellers, Holtzman, Bisk, Farhadi, and Choi}]{zellers2019hellaswag}
Rowan Zellers, Ari Holtzman, Yonatan Bisk, Ali Farhadi, and Yejin Choi. 2019.
\newblock Hellaswag: Can a machine really finish your sentence?
\newblock \emph{arXiv preprint arXiv:1905.07830}.

\bibitem[{Zeng et~al.(2023{\natexlab{a}})Zeng, Liu, Lu, Wang, Liu, Dong, and Tang}]{zeng2023agenttuning}
Aohan Zeng, Mingdao Liu, Rui Lu, Bowen Wang, Xiao Liu, Yuxiao Dong, and Jie Tang. 2023{\natexlab{a}}.
\newblock Agenttuning: Enabling generalized agent abilities for llms.
\newblock \emph{arXiv preprint arXiv:2310.12823}.

\bibitem[{Zeng et~al.(2023{\natexlab{b}})Zeng, Yu, Gao, Meng, Goyal, and Chen}]{zeng2023evaluating}
Zhiyuan Zeng, Jiatong Yu, Tianyu Gao, Yu~Meng, Tanya Goyal, and Danqi Chen. 2023{\natexlab{b}}.
\newblock Evaluating large language models at evaluating instruction following.
\newblock \emph{arXiv preprint arXiv:2310.07641}.

\bibitem[{Zhang et~al.(2023)Zhang, Singh, Liu, Liu, Yu, Gao, and Zhao}]{zhang2023tell}
Qingru Zhang, Chandan Singh, Liyuan Liu, Xiaodong Liu, Bin Yu, Jianfeng Gao, and Tuo Zhao. 2023.
\newblock Tell your model where to attend: Post-hoc attention steering for llms.
\newblock \emph{arXiv preprint arXiv:2311.02262}.

\bibitem[{Zheng et~al.(2024)Zheng, Chiang, Sheng, Zhuang, Wu, Zhuang, Lin, Li, Li, Xing et~al.}]{zheng2024judging}
Lianmin Zheng, Wei-Lin Chiang, Ying Sheng, Siyuan Zhuang, Zhanghao Wu, Yonghao Zhuang, Zi~Lin, Zhuohan Li, Dacheng Li, Eric Xing, et~al. 2024.
\newblock Judging llm-as-a-judge with mt-bench and chatbot arena.
\newblock \emph{Advances in Neural Information Processing Systems}, 36.

\bibitem[{Zhou et~al.(2024)Zhou, Liu, Xu, Iyer, Sun, Mao, Ma, Efrat, Yu, Yu et~al.}]{zhou2024lima}
Chunting Zhou, Pengfei Liu, Puxin Xu, Srinivasan Iyer, Jiao Sun, Yuning Mao, Xuezhe Ma, Avia Efrat, Ping Yu, Lili Yu, et~al. 2024.
\newblock Lima: Less is more for alignment.
\newblock \emph{Advances in Neural Information Processing Systems}, 36.

\bibitem[{Zhou et~al.(2023{\natexlab{a}})Zhou, Lu, Mishra, Brahma, Basu, Luan, Zhou, and Hou}]{zhou2023instruction}
Jeffrey Zhou, Tianjian Lu, Swaroop Mishra, Siddhartha Brahma, Sujoy Basu, Yi~Luan, Denny Zhou, and Le~Hou. 2023{\natexlab{a}}.
\newblock Instruction-following evaluation for large language models.
\newblock \emph{arXiv preprint arXiv:2311.07911}.

\bibitem[{Zhou et~al.(2023{\natexlab{b}})Zhou, Jiang, Wilcox, Cotterell, and Sachan}]{zhou2023controlled}
Wangchunshu Zhou, Yuchen~Eleanor Jiang, Ethan Wilcox, Ryan Cotterell, and Mrinmaya Sachan. 2023{\natexlab{b}}.
\newblock Controlled text generation with natural language instructions.
\newblock In \emph{International Conference on Machine Learning}, pages 42602--42613. PMLR.

\end{thebibliography}
\clearpage

\appendix

\newcolumntype{d}{>{\columncolor{brown!10}}r}
\newcolumntype{q}{>{\columncolor{Green!10}}r}

\setlength\tabcolsep{2pt}
\begin{table}[t]
  \centering
  \small
    \resizebox{1\linewidth}{!}{
\begin{tabular}{llcccccbcccccb}
\toprule
\multirow{2}{*}{Benchmark} & \multirow{2}{*}{Type} & \multicolumn{6}{c}{\textbf{Training Set}} & \multicolumn{6}{c}{\textbf{Test Set}} \\
\cmidrule(lr){3-8} \cmidrule(lr){9-14}
& & \textbf{L1} & \textbf{L2} & \textbf{L3} & \textbf{L4} & \textbf{L5} & \textbf{Avg.} & \textbf{L1} & \textbf{L2} & \textbf{L3} & \textbf{L4} & \textbf{L5} & \textbf{Avg.} \\
\midrule
\multirow{7}{*}{FollowBench} & Example & 31 & 20 & 17 & 16 & 16 & 100 & 9 & 20 & 23 & 24 & 24 & 100 \\
& Content & 16 & 15 & 17 & 15 & 12 & 75 & 9 & 10 & 8 & 10 & 13 & 50 \\
& Situation & 14 & 13 & 13 & 13 & 13 & 66 & 8 & 9 & 9 & 9 & 9 & 44 \\
& Style & 19 & 19 & 18 & 18 & 16 & 90 & 11 & 11 & 12 & 12 & 14 & 60 \\
& Format & 20 & 19 & 17 & 18 & 16 & 90 & 10 & 11 & 13 & 12 & 14 & 60 \\
& Mixed & 14 & 10 & 11 & 7 & 6 & 48 & 3 & 7 & 6 & 10 & 11 & 37 \\
& Total & 114 & 96 & 93 & 87 & 79 & 469 & 50 & 68 & 71 & 77 & 85 & 351 \\
\cdashlinelr{1-13}
\multirow{2}{*}{IFEval} & One-cons & - & - & - & - & - & 92 & - & - & - & - & - & 213 \\
& Multi-cons & - & - & - & - & - & 92 & - & - & - & - & - & 144 \\
\bottomrule
\end{tabular}
}
  \caption{The statistic of the datasets constructed in the empirical study.}
  \label{tab:100train_test}
\end{table}

\newcolumntype{b}{>{\columncolor{blue!10}}r}
\newcolumntype{d}{>{\columncolor{brown!10}}r}
\newcolumntype{q}{>{\columncolor{Green!10}}r}

\setlength\tabcolsep{6pt}
\begin{table}[!htb]
  \centering
    \resizebox{0.48\textwidth}{!}{
\begin{tabular}{lcccccb}

\toprule

\textbf{Data Selection Method}                                 & $C_1$ & $C_2$ & $C_3$ & $C_4$ & $C_5$ & \textbf{Total} \\
\midrule                    
    
All              & 61 & 54 & 431 & 493 & 428 & 1467  \\
Select & 60 & 48 & 192 & 136 & 79  & 515   \\
Random    & 19 & 19 & 143 & 178 & 156 & 515   \\
\bottomrule
\end{tabular}
}
  \caption{The statistic of the data used in \S\ref{sec:dataquality}. $C_i$ indicates that there are $i$ constraints in the instruction.
  }
  \label{tab:100quality_data}
\end{table}

\section{Appendix}
\label{sec:appendix}

\newcolumntype{b}{>{\columncolor{blue!10}}r}
\newcolumntype{d}{>{\columncolor{brown!10}}r}

\setlength\tabcolsep{0.5pt}
\begin{table}[t]
  \centering
  \scriptsize
    \resizebox{0.48\textwidth}{!}{
\begin{tabular}{lccccb}

\toprule

\textbf{Models} & \textbf{Format} & \textbf{Input} & \textbf{Task} & \textbf{Count} & \textbf{Average} \\

\midrule

LLaMA2-13B-Chat & 64.00                 & 34.20                & 28.00               & 67.40                      & 40.20            \\
$\text{Ours-LLaMA2-13B}_\text{Contrastive}$            & 54.90                 & 44.00                & 41.30               & 60.70                      & 44.20 \\
$\text{Ours-LLaMA2-13B}_\text{Discrimination}$            & 62.70                 & 49.00                & 55.00               & 71.30                      & 56.10            \\
\cdashlinelr{1-6}
GPT3.5-turbo    & 89.90                 & 76.00                & 79.90               & 70.00                      & 79.40            \\
GPT-4           & 91.10                 & 79.60                & 79.20               & 72.40                      & 82.20           \\
\bottomrule
\end{tabular}
}
  \caption{The overall performance of models on CELLO. \textit{Format}, \textit{Task}, \textit{Input}, \textit{Count} denote the criteria \textit{Answer format}, \textit{Task-prescribed phrases}, \textit{Input-dependent query}, and \textit{Count limit} respectively.
  }
  \label{tab:050cello}
\end{table}

\newcolumntype{b}{>{\columncolor{blue!10}}r}

\setlength\tabcolsep{1pt}
\begin{table}[!htb]
  \centering
    \resizebox{0.48\textwidth}{!}{
\begin{tabular}{lccccccb}

\toprule

\textbf{Models}      & \textbf{Content} & \textbf{Example} & \textbf{Format} & \textbf{Situation} & \textbf{Style} & \textbf{Mixed} & \textbf{Total} \\
\midrule
LLaMA2-13B-Chat      & 41.60            & 00.00             & 58.00           & 42.73              & 84.00          & 25.88          & 42.04          \\
$\text{Ours-LLaMA2-13B}_\text{Discrimination}$  & 40.80            & 05.00             & 58.67           & 37.27              & 74.00          & 38.82          & 42.43          \\
$\text{Ours-LLaMA2-13B}_\text{Contrastive}$ & 43.20            & 05.00             & 57.33           & 37.27              & 77.33          & 36.47          & 42.77          \\

\bottomrule
\end{tabular}
}
  \caption{Overall performance of models across different constraint categories on Followbench. 
  }
  \label{tab:100followbench}
\end{table}

\subsection{Details of Empirical Studies} \label{sec:appendix_1}

In \S\ref{sec:emp}, we first investigate what training data is effective in enhancing complex constraints following ability.
To achieve this, we split the instructions in the existing instruction following benchmarks, i.e., Followbench~\cite{jiang2023followbench} and IFEval~\cite{zhou2023instruction} into the training and test sets.
The training sets consist of two types of data:
(1) Compositional data: From IFEval, we utilize all the instructions with more than one constraint and all level-4 and level-5 instructions from Followbench.
(2) Atomic data: From IFEval, we use only one-constraint instructions. From Followbench, we use all level-1 and part of level-2 instructions to ensure an equal number of compositional and atomic data for fair comparison.

After collecting the instructions, we first employ GPT3.5-turbo to generate the answers to the corresponding instructions.
To improve the quality of the training data, we filter the samples from Followbench by prompting GPT3.5-turbo (We use the evaluation prompt from the original paper) and those from IFEval via its provided test scripts. 

The statistics of our training set and test set are provided in Tab.~\ref{tab:100train_test}.
It can be seen that there is a distribution shift between the training set and test set from FollowBench.
This may be because we use outputs satisfying all instruction constraints judged by GPT-3.5-turbo for training, with the rest as the test set. 
Consequently, the test set can be more challenging than the training data, especially for instructions with more constraints (level 4, level 5).
This can partially explain the results that training with compositional data boosts performance on instructions with 1 to 3 constraints but lowers it on those with 4 to 5 constraints.


\subsection{Complex Structure Synthesis}
As stated in \S\ref{sec:data}, we employ GPT-3.5-turbo to diversify the description for the same constraint. The corresponding prompt is shown in Tab.~\ref{tab:100diverse}. It is worth noting that, for the \textit{keyword} constraint, we prompt GPT3.5-turbo to brainstorm some keywords related to the instruction, shown in Tab.~\ref{tab:100keyword}. Then, we randomly select one of them and incorporate it into the diversified description to form the final instruciton, e.g., your response should not include the word ``architecture''.

\begin{table*}[t]
\small
    \begin{tabularx}{\linewidth}{X}
    \toprule
    \color{gray}{/* \textit{Task prompt} */}\\
    You are provided with a <constraint> in an instruction. As a prompt engineer, your task is to rephrase the provided <constraint> to make it more diverse. You ought to provide five more variants of the <constraint>. Make sure your revision does not change the meaning of the original <constraint>. \\
    \color{gray}{/* \textit{Example} */}\\
    ---INPUT--- \\
    <constraint>:\\
    Your response should contain at least 3 sentences.\\
    ---OUTPUT---\\
    variants:\\
    1. Respond with at least three sentences\\
    2. Use at least 3 sentences in your reply\\
    3. Your entire response should include at least three sentences\\
    4. Organize your entire response in at least 3 sentences\\
    5. Please make sure the response is at least 3 sentences long\\
    \color{gray}{/* \textit{Input} */}\\
    ---INPUT---\\
    <constraint>:\\
    \{\textbf{Given\_constraint}\}\\
    ---OUTPUT---\\
    variants:\\
    \bottomrule
    \end{tabularx}
  \caption{
    The prompts for diversifying the descriptions of a given constraint. We utilize one-shot in-context learning to enhance the performance. The information that requires manual input is highlighted in bold.
  }
  \label{tab:100diverse}
\end{table*}

\begin{table*}[t]
\small
    \begin{tabularx}{\linewidth}{X}
    \toprule
    \color{gray}{/* \textit{Task prompt} */}\\
    You are provided with an <instruction>. Your object is to come up some keywords that may be used to answer the <instruction>. They are usually related to the task described in the <instruction>. you should output your thinking process and the keywords you come up with. \\
    \color{gray}{/* \textit{Example} */}\\
    ---INPUT--- \\
    <instruction>:\\
    Explain Generative Adversarial Networks (GANs) to me using bullet points. Do not contain any commas in your response. \\
    ---OUTPUT---\\
    Thinking process:\\
    The <instruction> asks to explain GANs, hence, ``architecture'', ``training'' and ``generator'' may be appropriate keywords to use in the answer.\\
    Keywords:\\
    \textnormal{[}``architecture'', ``training'', ``generator'' \textnormal{]} \\
    \color{gray}{/* \textit{Input} */}\\
    ---INPUT---\\
    <instruction>:\\
    \{\textbf{Given\_instruction}\}\\
    ---OUTPUT--- \\
    \bottomrule
    \end{tabularx}
  \caption{
    The prompts for brainstorming some related keywords of a given instruction. The information that requires manual input is highlighted in bold.
  }
  \label{tab:100keyword}
\end{table*}

\begin{table*}[t]
\small
    \begin{tabularx}{\linewidth}{X}
    \toprule
    \color{gray}{/* \textit{Task prompt} */}\\
    You are provided with a response which is generated by a LLM and a constraint that the response is asked to follow. Now, you have known that the response does not follow the constraint. You are designated as a corrector to correct the response. You should make as minimal revisions as possible so that it follows the constraint. For example, you should not change the case of the word if you are not asked. To fulfil this task, you are expected to provide your analysis and a revised response which has followed the constraint.  \\
    \color{gray}{/* \textit{Example} */}\\
    ---INPUT---\\
    Response:\\
    <<Title>>: ISO Code for Andorra. The International Organization for Standardization (ISO) code for Andorra is <<ISO Code: 012>>. Andorra is a small, independent principality located in the Pyrenees mountains. The ISO code is a three-digit number that represents countries. I hope this information is helpful! Do you agree?\\
    Constraint: \\
    The very last sentence of your response should be ``Hope you agree with me.''\\
    ---OUTPUT--- \\
    Analysis: \\
    The last sentence of the response is ``Do you agree?''. I need to change it to ``Hope you agree with me.'' to follow the constraint.\\
    Revised response: \\
    <<Title>>: ISO Code for Andorra. The International Organization for Standardization (ISO) code for Andorra is <<ISO Code: 012>>. Andorra is a small, independent principality located in the Pyrenees mountain. The ISO code is a three-digit number that represents countries. I hope this information is helpful! Hope you agree with me.\\
    \color{gray}{/* \textit{Input} */}\\
    ---INPUT---\\
    Response:\\
    \{\textbf{Given\_response}\}\\
    Constraint:\\
    \{\textbf{Given\_constraint}\}\\
    ---OUTPUT---\\
    \bottomrule
    \end{tabularx}
  \caption{
    The prompts for correcting the response generated by the model to follow a specific constraint. The information that requires manual input is highlighted in bold.
  }
  \label{tab:100correct}
\end{table*}

\begin{table*}[t]
\small
    \begin{tabularx}{\linewidth}{X}
    \toprule
    \color{gray}{/* \textit{Prompt} */}\\
    You are a helpful assistant who reviews a debate among four other assistants in evaluating the quality of the outputs for a given instruction. The four assistants, Assistant (LLaMA2-13B-Chat), Assistant ($\text{Ours-LLaMA2-13B}_\text{Generation}$), Assistant ($\text{Ours-LLaMA2-13B}_\text{Discrimination}$ ), and Assistant ($\text{Ours-LLaMA2-13B}_\text{Contrastive}$), are given an instruction. Output (LLaMA2-13B-Chat), Output ($\text{Ours-LLaMA2-13B}_\text{Generation}$), Output ($\text{Ours-LLaMA2-13B}_\text{Discrimination}$ ), and Output ($\text{Ours-LLaMA2-13B}_\text{Contrastive}$) are generated by four different AI chatbots respectively. Assistants have conflicting evaluations. Your goal is to rate each output, assigning higher scores to the assistants whose responses better fulfill the given instruction.\\Here are some rules of the evaluation:\\1) You should prioritize evaluating whether the output honestly, precisely, and closely executes the instruction, then consider its helpfulness, accuracy, level of detail, harmlessness, etc.\\2) Outputs should NOT contain more or less than what the instruction asks for; as such outputs do NOT precisely execute the instruction.\\Rate each output from 1 to 10, then output your final verdict using this format: [[LLaMA2-13B-Chat-x]], [[$\text{Ours-LLaMA2-13B}_\text{Generation}$-x]], [[($\text{Ours-LLaMA2-13B}_\text{Discrimination}$)-x]], [[($\text{Ours-LLaMA2-13B}_\text{Contrastive}$)-x]], where x is the score you assigned to each assistant.\\
    \color{gray}{/* \textit{Instruction} */}\\
    \{\textbf{Given\_instruction}\}\\
    \color{gray}{/* \textit{Random order of four model outputs} */}\\
    <The Start of Assistant's Answer>\\
    \{\textbf{Model\_output}\}\\
    <The End of Assistant's Answer>\\
    \bottomrule
    \end{tabularx}
  \caption{
    The prompts for scoring the outputs generated by different models to a general instruction. The information that requires manual input is highlighted in bold.
  }
  \label{tab:100prompt_generalIF}
\end{table*}

\subsection{Generalization Experiments}
\subsubsection{General Instruction Following Ability}
\label{sec:appendix_general_instruction}
We adopt GPT-4 to compare and score the four candidates outputs given by LLaMA2-13B-Chat, $\text{Ours}_\text{Generation}$, $\text{Ours}_\text{Discrimination}$ and $\text{Ours}_\text{Contrastive}$ respectively.
The score ranges from 1 to 10.
To mitigate potential position bias in candidate outputs, we randomly shuffle the positions of the four candidate answers for each sample. 
The evaluation prompt is detailed in Tab.~\ref{tab:100prompt_generalIF}. 
Finally, we average the scores across all data samples.

\subsubsection{Out-of-Domain Generalization}
\label{sec:appendix_out_domain}

We provide detailed performance metrics and data examples for two out-of-domain complex instruction following benchmarks: FollowBench and CELLO.
The detailed performance of FollowBench is shown in Tab.~\ref{tab:100followbench}.
Except for mixed categories, our methods typically exhibit decreased performance compared to the backbone model when evaluated against individual, unseen constraints.
The declined performance in specific categories is reasonable. 
The complex instructions in specific categories (e.g., Style) from FollowBench (each has constraints from the same category) differ significantly from those in our training dataset (each contains constraints from multiple categories).
We show some cases in the Tab.~\ref{tab:100case}, with the constraints highlighted in bold.
This suggests that models training with certain constraints can hardly generalize to unseen constraints directly.
The detailed performance of CELLO is shown in Tab.~\ref{tab:050cello}.
As demonstrated in Tab.~\ref{tab:100case}, CELLO's constraints and language significantly differ from our training data.

\subsubsection{In-Domain Generalization} 
\label{sec:appendix_in_domain}
We detail the test set construction process below.
First, we select 200 instructions from the Open Assistant dataset (introduced in \S\ref{sec:data}) not in our training set. 
Next, we randomly choose 3 to 5 constraints from IFEval, pair them with descriptions from our description pool (\S\ref{sec:data}), and utilize GPT-3.5-turbo to paraphrase them, ensuring distinct descriptions from the training data. 
Additionally, we manually adjust specific requirements in the instructions, changing symbols (e.g., ``separated by 6 asterisk symbols ******'' to ``separate the responses with 6 hash signs: \#\#\#\#\#\#'') and formats (e.g., ``wrap the entire output in JSON format'' to ``I want the entire output in XML format'').

\subsubsection{Adversarial Setting}
\label{sec:appendix_in_adver}
We detail the test set construction process below.
Specifically, we utilize the same 200 seed instructions from \S\ref{sec:indomain} and the method introduced in \S\ref{sec:data} to append 6 to 7 constraints to the seed instructions.
These new instructions are challenging since our training data contains 3 to 5 constraints.





\subsection{Case Study}
We present some examples of various models following complex instructions in Tab.~\ref{tab:100more_case} and Tab.~\ref{tab:100more_case1}.
Given the complex instructions with multiple constraints, we present the outputs generated by GPT3.5-turbo and LLaMA2-13B-Chat and the list indicating whether they have followed the specific constraint or not.
Constraints in the instruction are \underline{underlined} sequentially.

\begin{table*}[t]
\small
    \begin{tabularx}{\linewidth}{X}
    \toprule
    \rowcolor[gray]{0.95}\multicolumn{1}{c}{\textbf{Instruction}} \\           
    Describe the content of the article in a brief manner. A study published earlier this year by Zee and her team examined the role of light in sleep for healthy adults in their 20s. Sleeping for only one night with a dim light, such as a TV set with the sound off, raised the blood sugar and heart rate of the young people during the sleep lab experiment. An elevated heart rate at night has been shown in prior studies to be a risk factor for future heart disease and early death, while higher blood sugar levels are a sign of insulin resistance, which can ultimately lead to type 2 diabetes. \underline{(1) The answer should be in at least \textbf{4} sections with each section starting with section X (where X is 1, 2, 3,...)}. The answer should be \underline{(2) written in English, with all letters lowercased}. Respond with \underline{(3) less than \textbf{73} words}. \\
    \rowcolor[gray]{0.95}\multicolumn{1}{c}{\textbf{GPT3.5-turbo}} \\
    \color{gray}{/* \textit{Output} */}\\
    section 1: study by zee and team on light's impact on sleep for young adults\\section 2: sleeping with dim light raised blood sugar and heart rate\\section 3: higher blood sugar levels indicate insulin resistance and risk of type 2 diabetes.\\
    \color{gray}{/* \textit{Followed list} */}\\
    \textnormal{[}False, True, True\textnormal{]} \\
    \rowcolor[gray]{0.95}\multicolumn{1}{c}{\textbf{ LLaMA2-13B-Chat}} \\
    \color{gray}{/* \textit{Output} */}\\
    section 1: \textbf{I}ntroduction\\the article discusses the impact of light on sleep and its effects on healthy adults in their 20s.\\section 2: \textbf{E}xperiment \textbf{F}indings\\the study found that sleeping with a dim light, such as a \textbf{TV} with the sound off, raised blood sugar and heart rate levels during the sleep lab experiment.\\section 3: \textbf{R}isk \textbf{F}actors\\elevated heart rate at night has been linked to future heart disease and early death, while higher blood sugar levels can lead to insulin resistance and type 2 diabetes.\\section 4: \textbf{C}onclusion\\the study highlights the importance of a dark, quiet sleep environment for optimal sleep and overall health.\\
    \color{gray}{/* \textit{Followed list} */}\\
    \textnormal{[}True, False, False\textnormal{]} \\
    \bottomrule
    
    \end{tabularx}
  \caption{
    The complex instructions and corresponding results of GPT3.5-turbo and LLaMA2-13B-Chat (Case 1).
  }
  \label{tab:100more_case}
\end{table*}

\begin{table*}[t]
\small
    \begin{tabularx}{\linewidth}{X}
    \toprule
    \rowcolor[gray]{0.95}\multicolumn{1}{c}{\textbf{Instruction}} \\           
    Summarize a meeting from the given list of bullet points. Be sure to convert shorthand into a first-hand account. Rose: Analyze data and presents findings\\John: propose new idea\\Jane: appointed to head project\\Tom: need more time to fix a software bug.\\ \underline{(1) End it with a postscript starting with P.S.}. \underline{(2) The \textbf{very end} of your entire response should be like: That is all you need!}. Your answer must also \underline{(3) contain at least \textbf{2} placeholders} (an example of a placeholder is [address]). \\
    \rowcolor[gray]{0.95}\multicolumn{1}{c}{\textbf{GPT3.5-turbo}} \\
    \color{gray}{/* \textit{Output} */}\\
    During the meeting, Rose analyzed data and presented her findings. John proposed a new idea for the project, while Jane was appointed to head the project. Tom mentioned that he needed more time to fix a software bug. \\P.S. \textnormal{[}Address\textnormal{]} That is all you need!\\
    \color{gray}{/* \textit{Followed list} */}\\
    \textnormal{[}True, True, False\textnormal{]} \\
    \rowcolor[gray]{0.95}\multicolumn{1}{c}{\textbf{ LLaMA2-13B-Chat}} \\
    \color{gray}{/* \textit{Output} */}\\
    Sure, I'd be happy to help! Here's a summary of the meeting based on the provided bullet points:\\The meeting began with Rose presenting her findings after analyzing the data. She shared some insightful observations and conclusions, which were well-received by the team.\\Next, John proposed a new idea that he believed would benefit the project. He explained his reasoning and provided some supporting evidence for his proposal.\\After John's proposal, Jane was appointed as the head of the project. She expressed her gratitude for the opportunity and outlined her vision for the project's success.\\Tom then reported that he needed more time to fix a software bug. He explained that the bug was more complex than initially thought and required additional time to resolve.\\In conclusion, the meeting was productive and successful. The team made progress on several fronts, including data analysis, proposal development, and project leadership.\\ \textbf{P.S. That is all you need!}\\\textnormal{[}Address\textnormal{]} - the physical location of the meeting \textnormal{[}Data\textnormal{]} - the specific data that Rose analyzed\\ \textnormal{[}Software bug\textnormal{]} - the technical issue that Tom is working to resolve\\
    \color{gray}{/* \textit{Followed list} */}\\
    \textnormal{[}True, False, True\textnormal{]} \\
    \bottomrule
    \end{tabularx}
  \caption{
    The complex instructions and corresponding results of GPT3.5-turbo and LLaMA2-13B-Chat (Case 2).
  }
  \label{tab:100more_case1}
\end{table*}

\subsection{Implementation Details}

We utilize 2 NVIDIA A800 80GB GPUs to conduct all the experiments. DeepSpeed ZeRO stage 1 is adopted for both SFT and DPO training. We use LORA\cite{hu2021lora} adaptor for effective training. We set the batch size to 4 for each GPU. 
All the methods utilizing SFT are trained for 2 epochs with the initial learning rate set to 3e-5 and the gradient accumulation step set to 4. The warm-up steps are set to 300. 
As for the training of methods utilizing DPO, the beta value is set to 0.1.
DPO is trained for 2 epochs with the initial learning rate set to 5e-4 and the gradient accumulation step also set to 4. 
We apply cosine learning rate scheduler and Adam optimizer to both models, and their maximum sequence length is set to 2048.

\newcolumntype{d}{>{\columncolor{brown!10}}r}
\newcolumntype{q}{>{\columncolor{Green!10}}r}

\setlength\tabcolsep{1pt}
\begin{table*}[!htb]
  \centering
  \scriptsize
    \resizebox{1\textwidth}{!}{
\begin{tabular}{lccccccccccdb}

\toprule

\textbf{Models} & \textbf{BaseModel} & \textbf{ChangeCase} & \textbf{Combination} & \textbf{Content} & \textbf{Format} & \textbf{Keywords} & \textbf{Language} & \textbf{Length} & \textbf{Punctuation} & \textbf{Startend} & \textbf{I-level} & \textbf{C-level} \\

\midrule
PaLM2-S                    &PaLM       & N/A   & N/A   & N/A   & N/A   & N/A   & N/A    & N/A   & N/A   & N/A   & 46.95 & 59.11 \\
GPT3.5-turbo               &GPT       & 66.29 & 75.38 & 88.68 & 89.17 & 74.23 & 100.00 & 65.03 & 24.24 & 86.57 & 63.96 & 73.62 \\
GPT4                       &GPT       & 78.65 & 72.31 & 96.23 & 94.27 & 88.34 & 96.77  & 76.92 & 69.70 & 95.52 & 78.74 & 85.13 \\
\cdashlinelr{1-13}
ChatGLM3-6B                &ChatGLM       & 16.85 & 21.54 & 67.92 & 45.86 & 56.44 & 54.84  & 38.46 & 34.85 & 56.72 & 30.31 & 43.41 \\
Qwen-14B-Chat              &Qwen       & 58.43 & 23.08 & 75.47 & 58.60 & 60.12 & 83.87  & 36.36 & 25.76 & 74.63 & 40.11 & 53.00 \\
LLaMA2-7B-Chat             &LLaMA2      & 47.19 & 12.31 & 79.25 & 58.60 & 62.58 & 29.03  & 43.36 & 16.67 & 56.72 & 36.60 & 48.68 \\
LLaMA2-13B-Chat            &LLaMA2       & 51.69 & 15.38 & 83.02 & 67.52 & 67.48 & 41.94  & 47.55 & 09.09 & 58.21 & 41.22 & 53.00 \\
LLaMA2-70B-Chat            &LLaMA2       & 49.44 & 27.69 & 79.25 & 65.61 & 72.39 & 22.58  & 48.25 & 21.21 & 70.15 & 43.44 & 55.40 \\
Vicuna-13B-V1.5            &LLaMA2       & 60.67 & 44.62 & 75.47 & 64.97 & 61.35 & 93.55  & 48.95 & 22.73 & 67.16 & 46.95 & 58.03 \\
WizardLM-13B-V1.2          &LLaMA2       & 57.30 & 21.54 & 75.47 & 70.70 & 70.55 & 93.55  & 55.94 & 25.76 & 71.64 & 49.72 & 60.55 \\
OpenChat-13B-V3.2          &LLaMA2       & 58.43 & 35.38 & 88.68 & 71.34 & 68.10 & 90.32  & 58.04 & 24.24 & 74.63 & 51.02 & 62.59 \\
Mistral-7B-Instruct-v0.2     &Mistral     & 68.54 & 26.15 & 88.68 & 77.71 & 77.30 & 80.65  & 56.64 & 27.27 & 79.10 & 56.19 & 65.95 \\
\cdashlinelr{1-13}
$\text{Ours-LLaMA2-7B}_\text{Generation}$ &LLaMA2   & 57.30 & 16.92 & 71.70 & 70.70 & 60.12 & 61.29 & 33.57 & 19.70 & 65.67 & 40.67 & 51.92 \\
$\text{Ours-LLaMA2-7B}_\text{Discrimination}$  &LLaMA2   & 55.06 & 09.23 & 77.36 & 64.97 & 61.35 & 74.19  & 40.56 & 21.21 & 79.10 & 43.99 & 53.48 \\
$\text{Ours-LLaMA2-13B}_\text{Generation}$    &LLaMA2    & 66.29 & 26.15 & 66.04 & 73.25 & 59.51 & 35.48  & 49.65 & 27.27 & 82.09 & 46.03 & 57.31 \\
$\text{Ours-LLaMA2-7B}_\text{Contrastive}$    &LLaMA2    & 77.53 & 15.38 & 75.47 & 70.70 & 55.83 & 67.74  & 46.85 & 31.82 & 89.55 & 46.95 & 58.75 \\
$\text{Ours-LLaMA2-13B}_\text{Discrimination}$ &LLaMA2   & 69.66 & 12.31 & 79.25 & 67.52 & 62.58 & 96.77  & 49.65 & 54.55 & 80.60 & 50.83 & 61.27 \\
$\text{Ours-LLaMA2-13B}_\text{Contrastive}$  &LLaMA2     & 69.66 & 16.92 & 84.91 & 68.15 & 66.87 & 93.55  & 51.05 & 57.58 & 88.06 & 52.13 & 63.91 \\
\cdashlinelr{1-13}
$\text{Ours-Mistral-7B}_\text{Generation}$      &Mistral     &76.40 & 50.77 & 66.04 & 78.98 & 61.35 & 58.06 & 55.94 & 46.97 & 92.54 & 54.90 & 66.07  \\
$\text{Ours-Mistral-7B}_\text{Contrastive}$    &Mistral       & 70.79 & 35.38 & 84.91 & 85.35 & 68.71 & 80.65  & 50.35 & 40.91 & 89.55 & 55.82 & 67.27 \\
$\text{Ours-Mistral-7B}_\text{Discrimination}$  &Mistral     & 82.02 & 20.00 & 71.70 & 81.53 & 63.19 & 96.77  & 55.24 & 62.12 & 85.07 & 56.75 & 67.39 \\

\bottomrule
\end{tabular}
}
\caption{The loose accuracy score (defined by \cite{zhou2023instruction}) of models on different constraints of the IFEval. To alleviate this false negative problem, following \cite{zhou2023instruction}, we use three variants of the model response to calculate a more loose accuracy score. 
  }
  \label{tab:100main}
\end{table*}

\begin{table*}[t]  
    \centering
    \scriptsize
    \resizebox{1\textwidth}{!}{
\begin{tabularx}{\textwidth}{lcX}
\toprule
\textbf{Source}   & \textbf{Category} & \multicolumn{1}{c}{\textbf{Instruction}} \\
\midrule
FollowBench       & Style             &  ..., \textbf{Position yourself as a sagacious detective}, ..., \textbf{Respond with the whimsical humor} and \textbf{imaginative wit} \textbf{typical of Lewis Carroll}, ..., placing emphasis on refined language and meticulous attention to detail \textbf{in a manner} \textbf{befitting the social and literary norms of the early 19th century.} \\
\midrule
FollowBench       & Mixed             & Lost, found vodka, drank to forget.\textbackslash n \textbackslash nAccording to the above prompt, write a \textbf{four-sentence} story that describes a man. However, \textbf{the word "man" should not appear} in the story. Please write \textbf{using an introspective narrative}  \textbf{tone}.You should also \textbf{describe something about the bad weather}.    \\
\midrule
CELLO             & Meta          & \begin{CJK}{UTF8}{gbsn}模仿以下格式，出一道题目："input": "一个物体质量2千克，以10米/秒的速度运动，它的动能是？" "output": "动能公式...\end{CJK} \\
\midrule
CELLO & Structure              & \begin{CJK}{UTF8}{gbsn}给定以下SQL文本，记录主键为f的薪水是多少？  ```  主键  性别  年龄  姓名  身高  a  女  70  q  213 ...  e  女  55  u  170  \end{CJK} \\
\midrule
Our Training data & N/A               & For the following request, please   \textbf{repeat the request itself} exactly as it is, ..., Create a list of subtopics for the given topic: Music theory. \textbf{Response should in all lowercase}. Your answer \textbf{must have a title} contained in double angular brackets, such as <<title>>. \textbf{Refrain from using commas} in your response. \\

\bottomrule
\end{tabularx}
}
  \caption{Data examples from FollowBench, CELLO and our training data.}
  \label{tab:100case}
\end{table*}

\subsection{Ablation Study on the Contrastive Method}
\label{sec:ablation_dpo}
To prove the effectiveness of the proposed objective function in \S\ref{sec:contrastive}, we train LLaMA2-13B-Chat only utilizing naive DPO loss $\mathcal{L}_{\text{DPO}}$. 
As shown in Tab.~\ref{tab:100dpo_only}, using only naive DPO loss causes models to output repeated constraints meaninglessly.
This is probably because DPO Loss tend to overfit to the training data~\cite{li2023policy, xu2024contrastive, hejna2023contrastive}.
After incorporating the $\mathcal{L}_{\text{SFT}}$ loss into our objective function, the model can effectively follow complex instructions.

\begin{table*}[t]
\small
    \begin{tabularx}{\linewidth}{X}
    \toprule
    \rowcolor[gray]{0.95}\multicolumn{1}{c}{\textbf{Instruction}} \\           
    Can you write a poem about the pros and cons of playing a lot of video games? Please make sure it's at least 40 sentences long (don't forget to add punctuation). You must highlight some words or phrases in your response, like *highlighted phrase*. \\
    \rowcolor[gray]{0.95}\multicolumn{1}{c}{\textbf{Output (w $\mathcal{L}_\text{DPO}$)}} \\
    *highlighted phrase* ``the pros and cons of playing a lot of video games''\\\\*highlighted phrase* ``the benefits and drawbacks of gaming''\\\\*highlighted phrase* ``the advantages and disadvantages of playing video games''\\\\*highlighted phrase* ``the positive and negative aspects of gaming''\\\\*highlighted phrase* ``the good and bad sides of gaming''\\\\*highlighted phrase* ``the pros and cons of gaming''\\\\*highlighted phrase* ``the benefits and drawbacks of gaming''...(repeated)\\
    \rowcolor[gray]{0.95}\multicolumn{1}{c}{\textbf{Output (w $\mathcal{L}_\text{Ours}$)}} \\
    Sure, I'd be happy to help! Here's a poem about the pros and cons of playing a lot of video games: Pros of playing video games:  * Fun and entertainment * Improved cognitive skills * Enhanced problem-solving abilities * Develops hand-eye coordination * Social connections with other gamers * Stress relief and relaxation * Escapism from real-life problems * Opportunities for creativity and self-expression  Cons of playing video games:  * Addiction and negative impact on mental health * Sleep deprivation and poor physical health * Social isolation and lack of face-to-face interaction * Increased aggression and violent behavior * Distraction from important responsibilities and tasks * Exposure to inappropriate content and negative influences * Wasting time and not being productive  It's important to note that while playing video games can have both positive and negative effects, the key is to find a balance and enjoy them responsibly. *Highlighted phrase:* It's crucial to be aware of the potential downsides and take steps to mitigate them, such as setting limits on playtime, engaging in physical activity, and prioritizing real-life relationships and responsibilities.\\
    \rowcolor[gray]{0.95}\multicolumn{1}{c}{\textbf{Instruction}} \\       
    Generate a forum thread about several people waiting to hear the latest local news. All sentences should be short. Refrain from using any commas. Use placeholders to represent different usernames. Use square brackets for placeholders, like \textnormal{[}username1\textnormal{]}, \textnormal{[}username2\textnormal{]}. Please include at least 20 placeholders in the thread. \\
    \rowcolor[gray]{0.95}\multicolumn{1}{c}{\textbf{Output (w $\mathcal{L}_\text{DPO}$)}} \\
    \textnormal \textnormal{[}username1\textnormal{]} \textnormal{[}username2\textnormal{]} \textnormal{[}username3\textnormal{]} \textnormal{[}username4\textnormal{]} \textnormal{[}username5\textnormal{]} \textnormal{[}username6\textnormal{]} \textnormal{[}username7\textnormal{]} \textnormal{[}username8\textnormal{]} \textnormal{[}username9\textnormal{]} \textnormal{[}username10\textnormal{]} \textnormal{[}username11\textnormal{]} \textnormal{[}username12\textnormal{]} ...(repeated)\\
    \rowcolor[gray]{0.95}\multicolumn{1}{c}{\textbf{Output (w $\mathcal{L}_\text{Ours}$)}} \\
    Title: Waiting for the Latest Local News  [username1]: Hey everyone, has anyone heard the latest local news? [username2]: No, have you? [username3]: Not yet, I'm waiting for the evening news. [username4]: Me too. I hope it's not more bad news. [username5]: I heard there might be a new development in the local project. [username6]: Really? Do you know anything about it? [username7]: No, I just heard it from a friend. [username8]: I hope it's something positive. We need some good news around here. [username9]: Agreed. I'm getting tired of all the negative news. [username10]: I know, it's been a tough few months. [username11]: I heard the local government is working on a new plan to improve the economy. [username12]: That sounds promising. Do you know what the plan is? [username13]: No, I don't have any details. But it's good to hear they're taking action. [username14]: Definitely. I'm looking forward to hearing more about it. [username15]: Me too. I hope it makes a difference. [username16]: I'm sure it will. The community needs something positive to focus on. [username17]: Absolutely. I'm feeling more hopeful already. [username18]: That's the spirit! We just need to stay positive and support each other. [username19]: Agreed. Let's keep each other updated on any news we hear. [username20]: Great idea! Let's keep the conversation going. \\
    \bottomrule
    \end{tabularx}
  \caption{
    The outputs generated by vanilla DPO loss and our training objective in \S\ref{sec:contrastive}.
  }
  \label{tab:100dpo_only}
\end{table*}





\end{document}